\documentclass[preprint,12pt,authoryear]{elsarticle}

\usepackage{amssymb}
\usepackage{array}
\usepackage{graphicx}
\usepackage{booktabs}
\usepackage{multirow}
\usepackage{adjustbox}
\usepackage{makecell}
\usepackage{enumitem}
\usepackage{caption}
\usepackage{amsmath}
\usepackage{adjustbox}
\usepackage{natbib}
\usepackage{hyperref}
\usepackage{csquotes}
\UseRawInputEncoding

\DeclareCaptionLabelFormat{myformat}{#2}
\journal{Engineering Applications of Artificial Intelligence }

\usepackage{graphicx} % 引入 graphicx 宏包
\usepackage{subcaption} % 引入 subcaption 宏包

\usepackage{amsmath} % 数学公式支持
\begin{document}

\begin{frontmatter}



\title{Inland Waterway Object Detection in Multi-environment: Dataset and Approach}

\author[label1,label2]{Shanshan Wang}
\author[label1,label3]{Haixiang Xu\corref{cor1}}
\author[label1,label3]{Hui Feng\corref{cor1}}
\cortext[cor1]{Co-corresponding author}
\author[label1,label3]{Xiaoqian Wang}
\author[label1,label3]{Pei Song}
\author[label1,label3]{Sijie Liu}
\author[label4]{Jianhua He}

\address[label1]{Key Laboratory of High Performance Ship Technology (Wuhan University of Technology), Ministry of Education, Wuhan, Hubei, China}
\address[label2]{College of Automobile Technology and Service, Wuhan City Polytechnic, Wuhan, Hubei, China}
\address[label3]{School of Naval Architecture, Ocean and Energy Power Engineering, Wuhan University of Technology, Wuhan, Hubei, China}
\address[label4]{School of Computer Science and Electronic Engineering, University of Essex, Colchester, CO4 3SQ, United Kingdom}

\begin{abstract}
The success of deep learning in the field of intelligent ship visual perception largely relies on information-rich image data. However, the dedicated datasets for inland waterway vessel objects remain scarce, failing to meet the adaptability requirements of visual perception systems in complex and dynamic environments. Particularly in inland waterway scenarios, due to narrow waterways, variable weather conditions, and interference from urban structures and lighting along the riverbanks, object detection systems based on existing inland waterway datasets exhibit significant limitations in robustness. To address these issues, this paper constructs a new vessel detection dataset named Multi-environment Inland Waterway Vessel Dataset (MEIWVD). The MEIWVD comprises 32,478 high-quality images from various inland waterway scenarios, covering complex environmental conditions such as sunny, rainy, foggy, and artificial lighting, etc. These images comprehensively encompass common vessel types in the Yangtze River Basin, while considering image diversity, sample independence, environmental complexity, and multi-scale characteristics, making MEIWVD a benchmark dataset with exceptional properties for vessel object detection. To leverage the characteristics of the MEIWVD, this paper proposes a scene-guided image enhancement module for multi-environment scenarios, which adaptively enhances water surface images based on environmental conditions to improve detector performance in complex scenarios. Additionally, a parameter-limited dilated convolution is introduced to enhance the representation of salient features of inland waterway vessels by leveraging their geometric characteristics. Finally, a multi-scale dilated residual fusion method is proposed to effectively integrate multi-scale features and improve the detection of multi-scale objects. Experimental results demonstrate that the MEIWVD dataset, constructed in this study, provides a more rigorous benchmark for object detection algorithms compared to other water surface object datasets, due to its broader range of scenarios. Furthermore, the proposed methods significantly improve the performance of object detectors, particularly in complex multi-environment dataset.
\end{abstract}


\end{frontmatter}

\section{Introduction}
\label{}
With the rapid development of artificial intelligence technology, intelligent ships and smart shipping have gradually become research hotspots in the field of waterway transportation, particularly in inland shipping and maritime supervision, where they are driving significant advancements. Inland shipping, as a vital link connecting cities and commerce, not only promotes regional economic development but also enhances logistics efficiency. Meanwhile, maritime supervision ensures the safety and compliance of shipping, effectively preventing environmental pollution and shipping accidents. To improve the safety of inland shipping and the intelligence of maritime supervision, it is urgent to rely on advanced intelligent perception technology, especially the vision-based water surface detection technology.

However, achieving precise and real-time detection of surface objects, including ships and buoys, continues to pose a significant challenge in ship perception across varied maritime environments. Surface object detection requires not only high precision and reliability but also consideration of the impact of different environmental conditions on sensor performance. Despite the revolutionary breakthroughs brought by the rapid progress of deep learning in the field of object detection, inland ship object detection still faces a series of technical challenges, among which the scarcity of datasets, limited scenario coverage, and the complex and variable weather conditions of inland waterways are particularly prominent. These issues directly limit the generalization ability and detection accuracy of deep learning models in practical applications.  

While a small number of studies have publicly released inland shipping datasets, the existing datasets are limited in quantity and incomplete in scenario coverage, especially lacking data under complex weather conditions \cite{shaoSeaShipsLargeScalePrecisely2018,zhengMcshipsLargescaleShip2020,wangMarineVesselDetection2024,iancuABOshipsInshoreOffshore2021,yangLightweightTheorydrivenNetwork2024}. Most datasets primarily focus on daytime scenarios under clear weather, with insufficient attention to ship detection in complex meteorological conditions such as cloudy, foggy, and rainy weather. This limitation not only affects the training effectiveness of models but also restricts their potential application  in real-world environments. Additionally, the authenticity of data is equally critical. Many datasets obtained through web crawlers significantly differ from the data distribution in actual usage scenarios. Due to this inconsistency, models struggle to generalize effectively to real-world scenarios during training, leading to diminished accuracy and reliability in object detection and recognition. In inland waterways, the uniqueness of the scenarios further exacerbates this challenge. For example, narrow waterways and complex backgrounds increase the difficulty of models adapting to practical application scenarios. Finally, the detection of long-distance small objects in inland environments poses higher demands, especially for small objects such as buoys, which are more challenging compared to large vessels like container ships and cargo ships. Therefore, it is critically important to construct a dataset that encompasses real-world scenarios, diverse environments, and multi-scale surface objects.

On the other hand, numerous studies have been conducted on the detection and recognition of water surface targets, such as \cite{guoRotationalLibraRCNN2020,caiFEYOLOYOLOShip2024,zhangHighPerformanceShip2022,xingSDETRTransformerModel2023}, the performance of these methods is often adversely affected in various complex scenarios, including rain, fog, and nighttime conditions, potentially leading to significant degradation in detection accuracy. Some studies, such as SharpGAN \cite{fengSharpGANDynamicScene2021} and D3-Net \cite{guoD3NetIntegrated2023}, have attempted to improve detection performance under adverse weather conditions through image enhancement techniques. However, due to the lack of multi-scenario water surface image datasets, existing research primarily relies on synthetic data to simulate images under rainy, foggy, and other challenging conditions. The discrepancy between synthetic and real-world data may result in insufficient generalization capability of the models, making it difficult to achieve satisfactory detection performance in practical complex scenarios. Therefore, developing more robust detection algorithms based on multi-scenario datasets is of both theoretical and practical significance.

To address the aforementioned challenges, we firstly propose a multi-environment inland waterway vessel dataset (MEIWVD), which is a comprehensive dataset for inland waterway vessel object detection and recognition. MEIWVD is constructed by collecting and organizing diverse image data from real-world inland waterway environments, covering various weather conditions and complex scenarios, thereby providing rich, diverse, and challenging training and testing data for deep learning models. Specifically, the dataset includes images under multiple weather conditions such as sunny, cloudy, foggy, and rainy conditions, as well as ship images under artificial lighting at night. Additionally, the dataset emphasizes the collection of multi-scale vessel information, with a high ratio of surface objects per image to align with real-world application needs. High-precision manual annotation ensures the accuracy and reliability of object detection. By constructing this benchmark dataset for multi-environment inland waterway vessel object detection, this paper aims to advance the application of object detection algorithms in complex inland environments, improving surface object detection and recognition accuracy and robustness.  

Additionally, to effectively address the challenges posed by the diverse scenarios in the MEIWVD, this paper proposes a series of innovative methods. Initially, in order to better deal with the complex multi-environment scene object detection, a scene-guided image enhancement (SGIE) method is introduced. This method uses scene-guided prompts to perform targeted image enhancement for different weather conditions, thereby improving the classification and detection performance of models in complex environments. Additionally, based on the geometric characteristics of surface objects, parameter-limited dilated convolution (PLD-Conv) is designed to enhance the model's ability to recognize the shapes of surface objects. Finally, a multi-scale dilated residual fusion (MS-DRF) module is proposed for multi-scale object detection. This module integrates information from different scales to enhance the model's detection performance for objects of various sizes. Through these methods, this paper aims to better adapt to the complex requirements of inland waterway vessel object detection, thereby improving the overall performance of the model.  

The main contributions of this paper are summarized as follows:   
\begin{enumerate}[label=(\arabic*)]
	\item A diverse and environmentally rich inland waterway dataset is constructed. The dataset covers a wide range of real-world scenarios in inland waterways, including common vessel categories (e.g., cargo ships, passenger ships, container ships) and various weather conditions (e.g., sunny, cloudy, foggy, etc.) as well as specific time conditions (e.g., daytime, post-dusk, and evening with artificial lighting). The comprehensiveness and diversity of the dataset provide rich resources for training and testing deep learning models, effectively reflecting the complexity of inland environments and improving the performance of object detectors.  
	\item To address the complexity and diversity of inland environments, this paper proposes scene-guided image enhancement (SGIE) module. By combining scenario-embedded vectors with guided prompts, the method is able to accurately model the degradation conditions, enabling targeted feature enhancement for different scenarios and improving the robustness and accuracy of detection models in multi-environment settings. Additionally, the method also demonstrates strong generalization capabilities in unseen degradation scenarios.  
	\item Given the relatively uniform shapes and fixed aspect ratios of surface objects, this paper proposes a parameter-limited dilated convolution (PLD-Conv) module. By designing different convolution strategies in horizontal and vertical directions, this module effectively captures the geometric features of surface objects, improving the model's performance in surface object detection and recognition.  
	\item To address the multi-scale object characteristics in the dataset, this paper designs a multi-scale dilated residual fusion (MS-DRF) module. This module efficiently captures multi-scale information from different receptive fields, for the purpose of enhancing feature representation, and reducing computational overhead. In addition, MS-DRF also effectively fuses multi-scale object features, improving the detector's ability to detect multi-scale objects.  
\end{enumerate}
The structure of this paper is organized as follows. Section 2 reviews related work on surface object detection datasets and state-of-the-art methods. In Section 3, we provide the details of the construction of the MEIWVD, including data collection, annotation methods, and data distribution analysis. Based on the characteristics of the MEIWVD, we introduce the SGIE, PLD-Conv, as well as MS-DRF module, elaborating on the principles and implementation of each method in Section 4. Section 5 conducts comprehensive benchmark testing and performance validation of the MEIWVD using typical detectors. At last, we summarize the research findings and discusses future research directions in Section 6.  

%\subsection{Subsection1.1}
%Text of Subsection1.1
%\subsubsection{Subsubsection1.1.1}
%This is the content of Subsubsection 1.1.1. daw d adwadawdna  dawdawdaw

\section{Related work}
\label{sec:sec2}
\subsection{Surface object datasets}
In the field of surface object detection, datasets are a cornerstone for research and algorithm development, with their scale, diversity, annotation accuracy, and real-world applicability being critical factors. Several widely used datasets have been applied to ship detection tasks, greatly advancing the field of surface object detection, but existing datasets have certain limitations that hinder their ability to fully capture the complexity of real-world environments. The SeaShips \cite{shaoSeaShipsLargeScalePrecisely2018} dataset, comprising 7,000 images across six ship categories, is commonly used in marine ship detection research, however, its limited data volume and diversity restrict its effectiveness in representing real-world scenarios. The SMD \cite{yangLightweightTheorydrivenNetwork2024} dataset, which provides visible and near-infrared images from 81 video clips, is suitable for multi-sensor data fusion tasks but lacks sufficient coverage of complex weather conditions. The McShips \cite{zhengMcshipsLargescaleShip2020} dataset, with 14,709 images, includes various military and civilian ship categories, reflecting some diversity in ship types, nevertheless, its high proportion of military ships deviates significantly from typical inland scenarios, limiting its practical applicability. In contrast, the ABOships \cite{iancuABOshipsInshoreOffshore2021} dataset covers nine object categories and investigates the impact of object size on detection accuracy, yet it still struggles with the challenge of small object detection, a persistent issue in the field. The recently proposed MVDD13 \cite{wangMarineVesselDetection2024} dataset consists of 35,474 images covering 13 ship categories and attempts to mirror the proportional distribution of ships in real-world scenarios. However, part of its data is collected through web crawling, and the exact proportion of such data is unknown. The diversity of data sources may compromise the overall balance of the dataset. 

Through the analysis of existing literature, while these datasets have advanced research in ship detection, their limitations in scale, diversity, and real-world applicability highlight the need for more comprehensive datasets to address the complexities of surface object detection. To address this gap, we constructed a large-scale inland waterway vessel dataset, MEIWVD, based on data collected from the Yangtze River Basin. 

\subsection{Surface object detection methods}
Vision-based surface object detection technology is one of the key technologies for intelligent ship perception, primarily used for real-time perception of surrounding environmental information to assist in autonomous navigation, collision avoidance, and automated maritime supervision. Traditional object detection methods \cite{loweDistinctiveImageFeatures2004,dalalHistogramsOrientedGradients2005} typically require manually designed extractor to extract the feature from the image, followed by in-depth analysis and recognition of these features using machine learning classifiers. Although these methods have achieved significant success in specific scenarios, due to manually designed features, their inherent limitations cannot be ignored, such as relatively limited generalization capabilities and high dependence on the designer's subjective judgment, which restricts their application in broader and more complex scenarios. Girshick et al. \cite{girshickRegionbasedConvolutionalNetworks2016} proposed the region-based convolutional neural network (R-CNN) for object detection, marking the entry of object detection technology into the era of deep learning. Generally, deep learning-based object detection can be divided into two categories, namely two-stage detection methods and single-stage detection methods. The former involves a coarse-to-fine process for proposal region generation, while the latter directly predicts detection boxes without a screening step. In two-stage detection methods, He et al. \cite{heSpatialPyramidPooling2015} proposed the spatial pyramid pooling network (SPPNet), which generates fixed-length feature representations. Fast RCNN \cite{girshickFastRCNN2015a} and Faster RCNN \cite{renFasterRCNNRealtime2017} further improved upon R-CNN and SPPNet by introducing the region proposal network (RPN), enabling the prediction of detection boxes and classification within a single network structure. Lin et al. \cite{linFeaturePyramidNetworks2017} proposed the feature pyramid network (FPN), a top-down architecture with lateral connections for building high-level semantics at all levels.  

Joseph et al. \cite{redmonYouOnlyLook2016} proposed YOLO (You Only Look Once) which is a classic single-stage detector. The YOLO series has since undergone continuous improvements, with subsequent releases including YOLOv2, YOLOv3, YOLOv5, YOLOv8, YOLOX, and YOLOv11 \cite{redmonYOLOv3IncrementalImprovement2018,bochkovskiyYOLOv4OptimalSpeed2020,mengYOLOv5sFogImprovedModel2023,wangYOLOv7TrainableBagofFreebies2023}. Wang et al. \cite{wangGoldYOLOEfficientObject2023} proposed GOLD-YOLO based on YOLO, enhancing multi-scale feature fusion capabilities and achieving a balance between speed and accuracy across different model scales. Compared to two-stage algorithms, the YOLO series emphasizes computational efficiency, offering faster inference speeds while maintaining detection accuracy, making YOLO the preferred choice for real-time object detection in robotics, autonomous driving, and video surveillance applications \cite{tervenComprehensiveReviewYOLO2023}. Liu et al. \cite{liuSSDSingleShot2016} proposed the single shot detector (SSD), introducing multi-reference and multi-resolution detection techniques to improve the accuracy of single-stage detectors. Lin et al. \cite{linFocalLossDense2020} proposed RetinaNet, which introduced focal loss to make the detector focus more on hard-to-classify samples during training, enabling single-stage detectors to achieve accuracy comparable to two-stage detectors while maintaining detection speed. Meng et al. \cite{xuEndtoEndSemiSupervisedObject2021} proposed an end-to-end semi-supervised object detection algorithm, using a soft teacher model to evaluate region proposals generated by a student model, reducing the need for data annotation.  

With the tremendous success of transformers \cite{vaswaniAttentionAllYou2017a} in natural language processing, some scholars have introduced self-attention mechanisms to address object detection problems. DETR \cite{carionEndtoEndObjectDetection2020a} was the first algorithm to introduce transformers into the field of object detection. Subsequently, to address the excessive computational cost of self-attention in image domains, Deformable DETR \cite{zhuDeformableDetrDeformable2021} drew inspiration from deformable convolutions \cite{daiDeformableConvolutionalNetworks2017a,zhuDeformableConvNetsV22019} and designed a deformable attention mechanism \cite{xiaVisionTransformerDeformable2022}, significantly improving model convergence speed. CF-DETR \cite{caoCFDETRCoarsetoFineTransformers2022} introduced a coarse-to-fine decoder layer to further enhance object detection accuracy. Subsequently, Cascade-DETR \cite{yeCascadeDETRDelvingHighQuality2023}, Pyramid Vision Transformer \cite{wangPyramidVisionTransformer2021}, and Rand-DETR \cite{puRankDETRHighQuality2023} introduced cascade, pyramid, and hierarchical architecture modules into transformer-based object detection mechanisms. Rekavandi et al. \cite{rekavandiTransformersSmallObject2023} analyzed the advantages of transformers in small object detection. Arkin et al. \cite{arkinSurveyObjectDetection2023} conducted a comparative analysis of CNN-based and Transformer-based object detection algorithms, highlighting their respective strengths and limitations. 

In the field of intelligent ships, YOLO series algorithms are often modified for surface object detection to ensure real-time performance. Meng et al. \cite{erShipDetectionDeep2023} analyzed the lag in the development of surface object detection compared to general object detection, attributing it to the lack of widely recognized benchmark datasets. They also analyzed existing surface object detection algorithms, pointing out that the research challenges lie in small object detection and interference from complex backgrounds and weather conditions. Guo et al. constructed D3-Net \cite{guoD3NetIntegrated2023}, which integrates dehazing, deblurring, and object detection tasks within a single network structure. Wang et al. \cite{wangNavigatingWatersObject2024} performed a systematic and in-depth evaluation of state-of-the-art real-time object detection algorithms, specifically focusing on their applicability to autonomous surface vehicles (ASVs). By adding 15 different types of distortions to the dataset, such as noise, blurring, fog, and contrast changes, they concluded that existing real-time object detection methods lack robustness under these weather variations.  

Deep learning-based object detection algorithms have made significant progress in recent years and have been widely applied in various fields. These algorithms automatically identify object objects in images or videos using deep neural networks and output their location and category information, offering advantages such as efficiency, accuracy, and robustness. In practical applications, selecting or designing appropriate object detection algorithms based on specific needs and scenarios is particularly important. 

\section{Multi-environment inland waterway vessels dataset}
\label{sec:sec3}
In this section, we present a comprehensive description of the construction process of the multi-environment inland waterway vessel dataset (MEIWVD). Specifically designed for ship object detection, the dataset comprises 32,478 images and covers four common inland surface object categories: cargo ships, passenger ships, buoys, and container ships. Although MEIWVD may not match certain marine datasets in terms of category diversity, it offers unique value in two critical aspects. First, it excels in the diversity of environmental scenarios, encompassing real inland scenes under various conditions such as daytime, post-dusk, rainy, foggy, and artificial lighting at night. Second, unlike marine ship datasets, inland vessel datasets are characterized by a distinct composition of targets, primarily featuring massive cargo ships, small passenger ships, and tiny buoys, each posing unique detection challenges due to their extreme size variations and specific operational contexts.

To begin with, we systematically elaborate on the dataset construction methods, focusing on data collection, annotation, and preprocessing, while highlighting the key features and advantages of the dataset. Furthermore, we provide fundamental statistical information about the dataset, including the number of images, category distribution, and weather conditions, to assist researchers in better understanding and utilizing this resource.

\subsection{Data collection and annotation}
\label{sec:sec3.1}
The MEIWVD was primarily constructed through data collection of common surface objects in inland waterways of the Yangtze River Basin. To achieve this, we installed Hikvision cameras on the riverbanks of the Yangtze River, adjusting their positions and angles to ensure comprehensive coverage of the water surface areas. However, not all raw images could be directly used for research. Based on our preprocessing strategy, a combination of automatic and manual methods was employed to refine the candidate images, as detailed below:  

\begin{enumerate}[label=(\arabic*)]
	\item Elimination of unrecognizable images: In adverse weather conditions, such as heavy rain or dense fog, the captured images exhibited severely compromised visibility due to raindrop occlusion or limited light penetration. These conditions rendered the images indistinguishable even to human observers, necessitating their exclusion from further analysis.
	\item Multi-environment data collection: To effectively cover common inland weather and lighting conditions, data from special weather conditions were exclusively processed and collected, ensuring accurate and precise annotation. To more realistically simulate the inland waterway environment and enhance detector training, we intentionally included images that are difficult for the human eye to recognize but can be accurately annotated using contextual information. This approach aims to strengthen the detector's ability to extract effective features from the data, significantly improving its performance in detecting surface objects, thereby providing more precise and efficient support in relevant application fields.
\end{enumerate}

\begin{figure}[htbp]
    \centering
    % 第一行
    \begin{subfigure}{0.24\textwidth}
        \includegraphics[width=\textwidth]{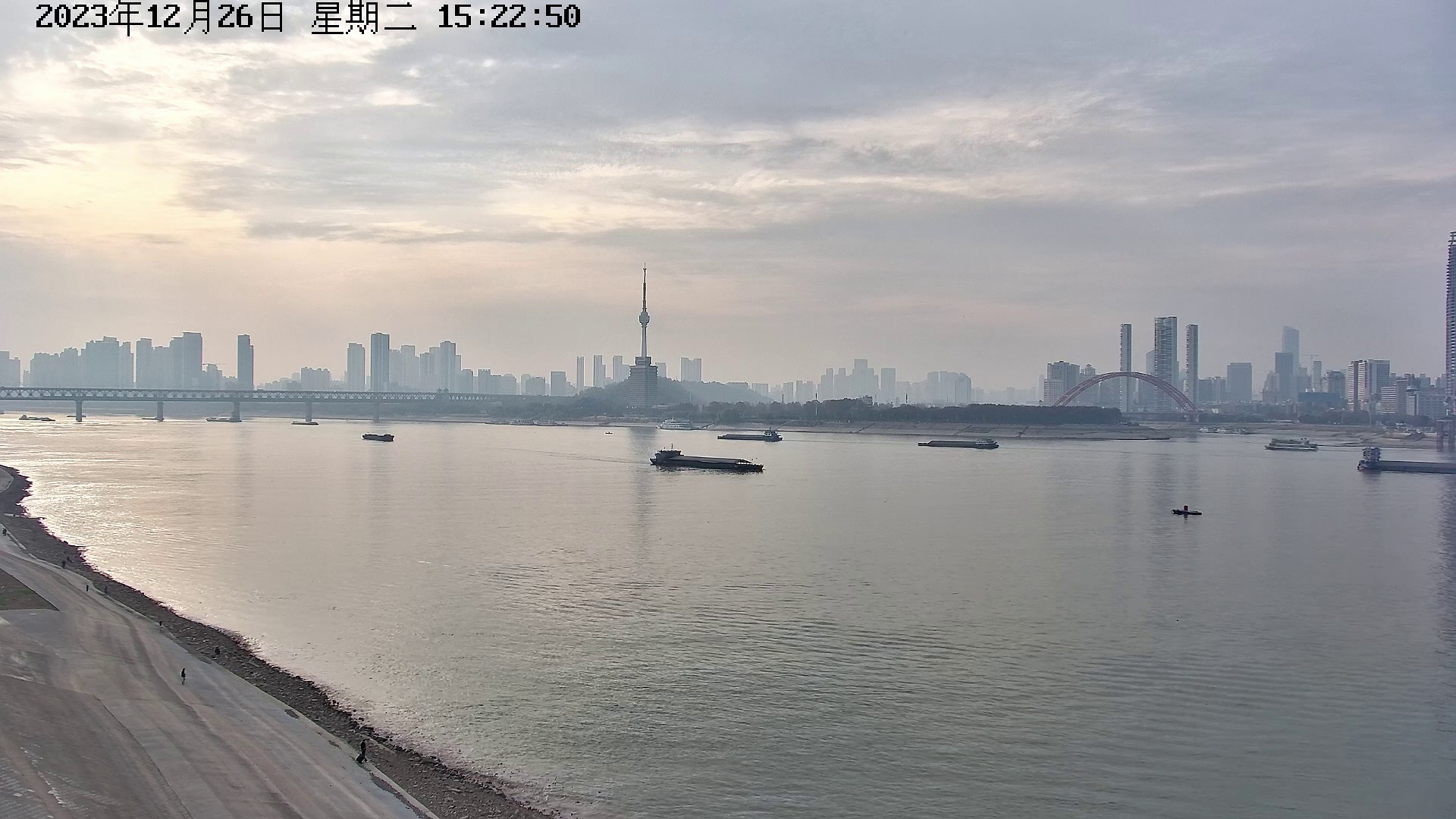}
        \label{fig:multi_view_1}
    \end{subfigure}
    \hfill
    \begin{subfigure}{0.24\textwidth}
        \includegraphics[width=\textwidth]{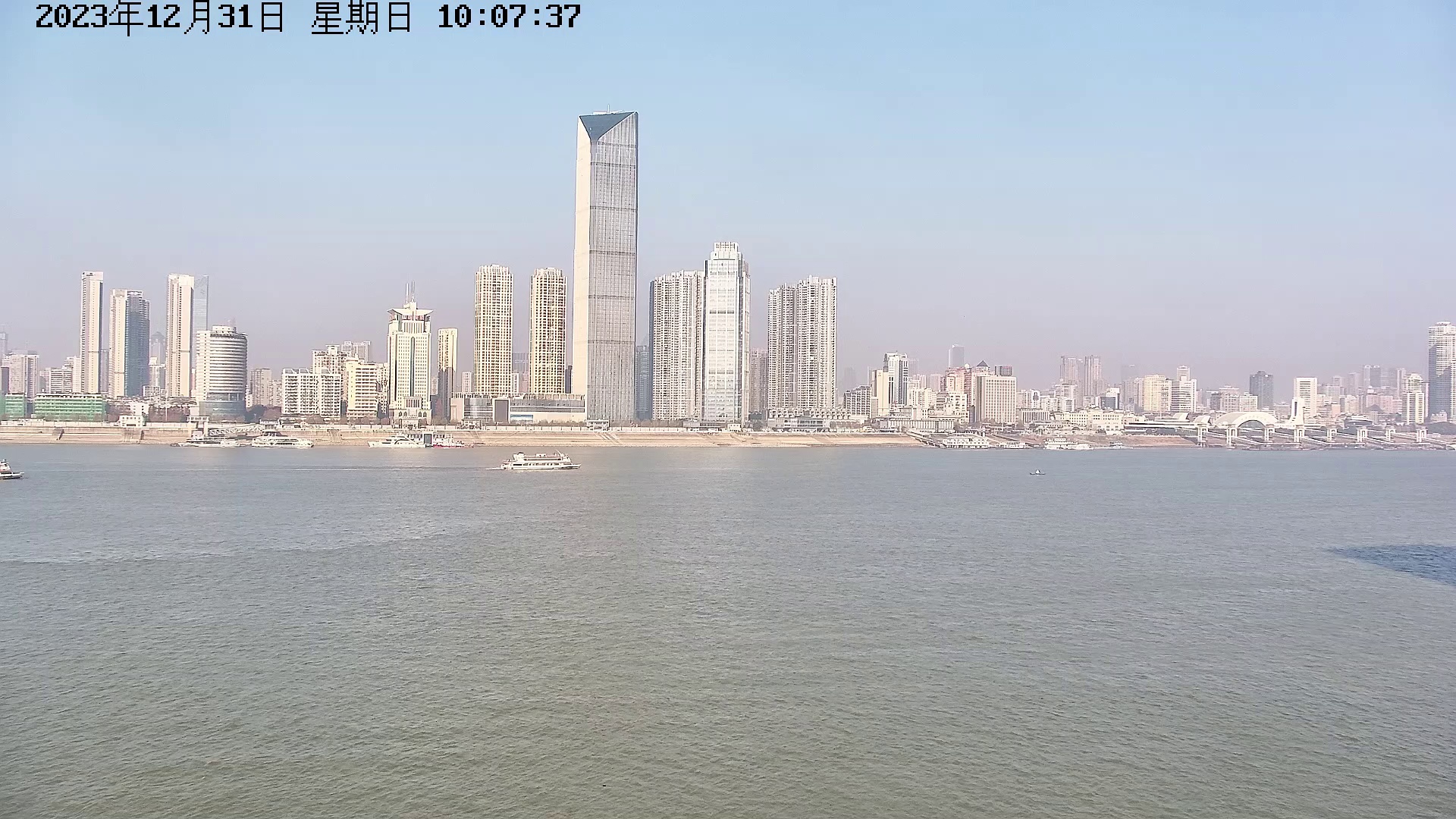}
        \label{fig:multi_light_1}
    \end{subfigure}
    \hfill
    \begin{subfigure}{0.24\textwidth}
        \includegraphics[width=\textwidth]{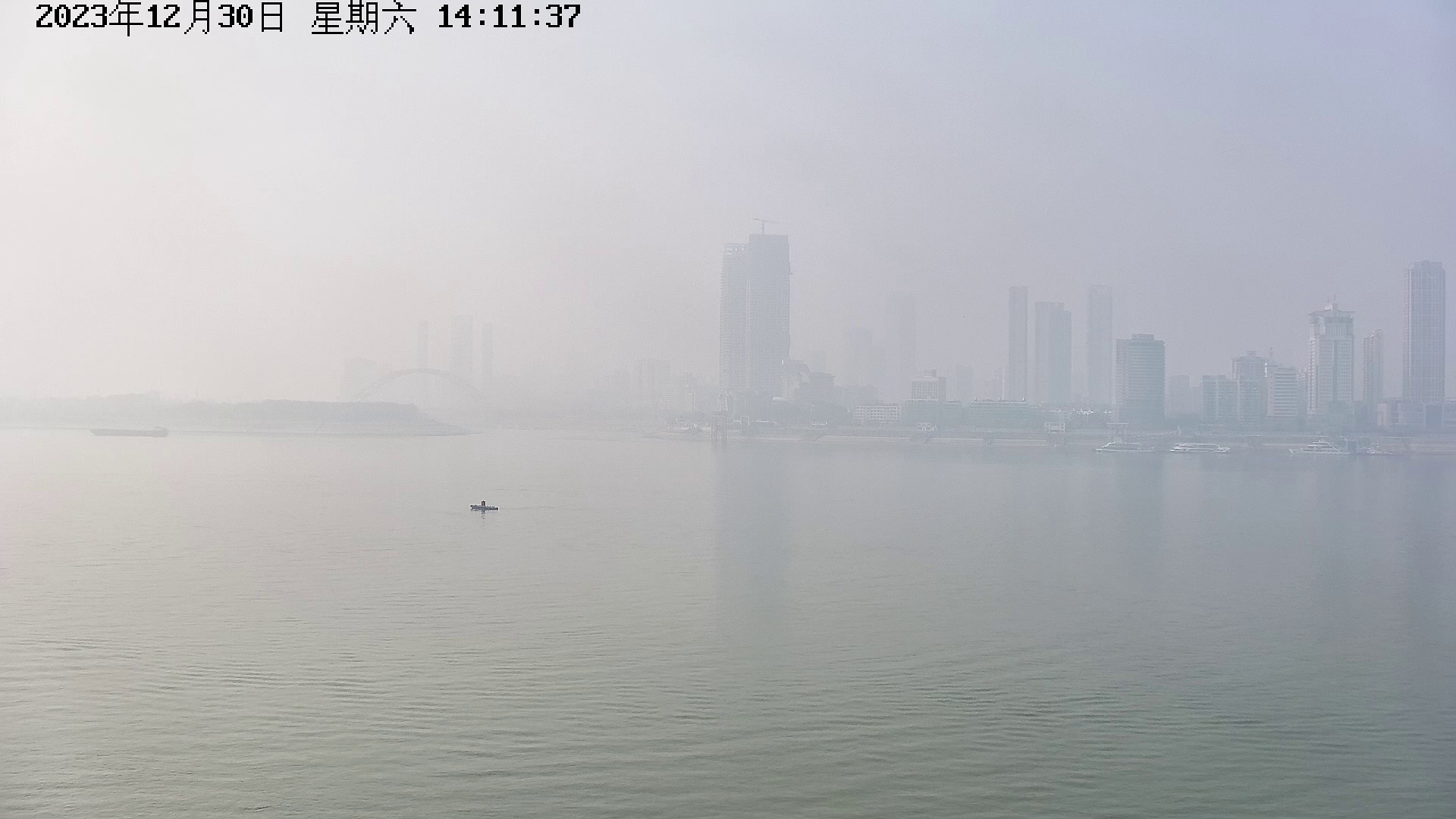}
        \label{fig:multi_weather_1}
    \end{subfigure}
    \hfill
    \begin{subfigure}{0.24\textwidth}
        \includegraphics[width=\textwidth]{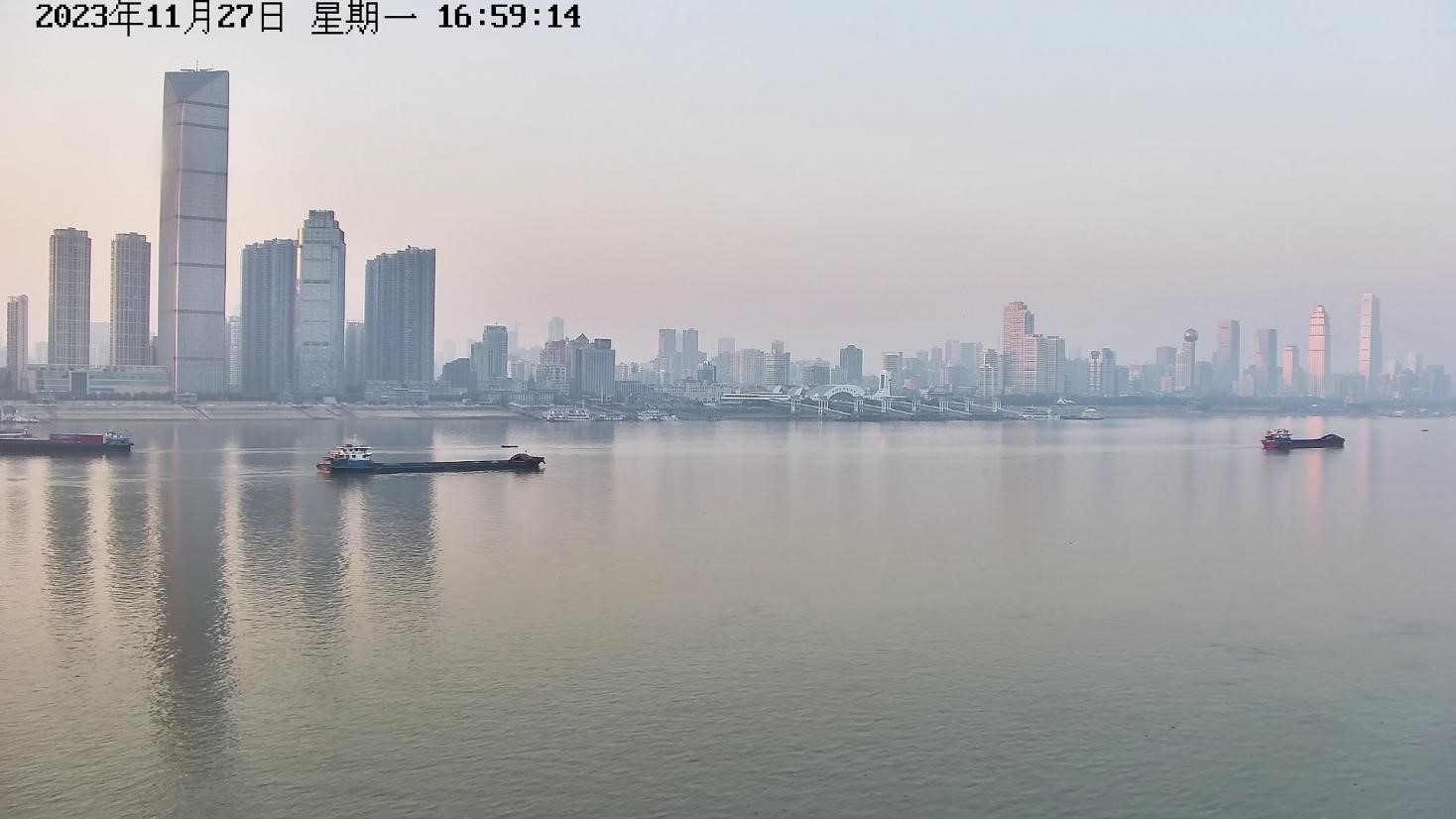}
        \label{fig:multi_scenario_1}
    \end{subfigure}
    \\ % 换行
    \vspace{-1em}
    % 第二行
    \begin{subfigure}{0.24\textwidth}
        \includegraphics[width=\textwidth]{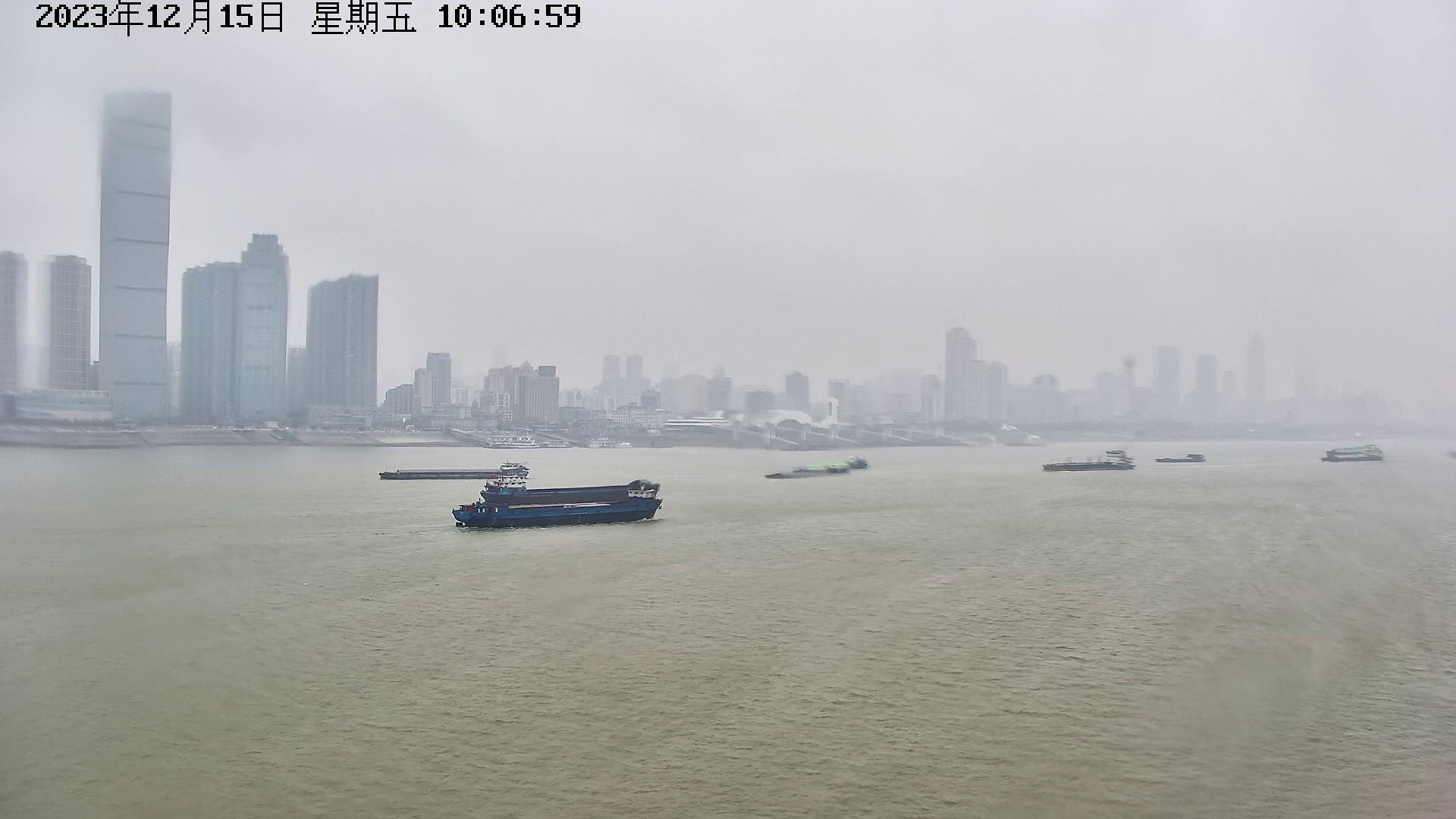}
        \label{fig:multi_view_2}
    \end{subfigure}
    \hfill
    \begin{subfigure}{0.24\textwidth}
        \includegraphics[width=\textwidth]{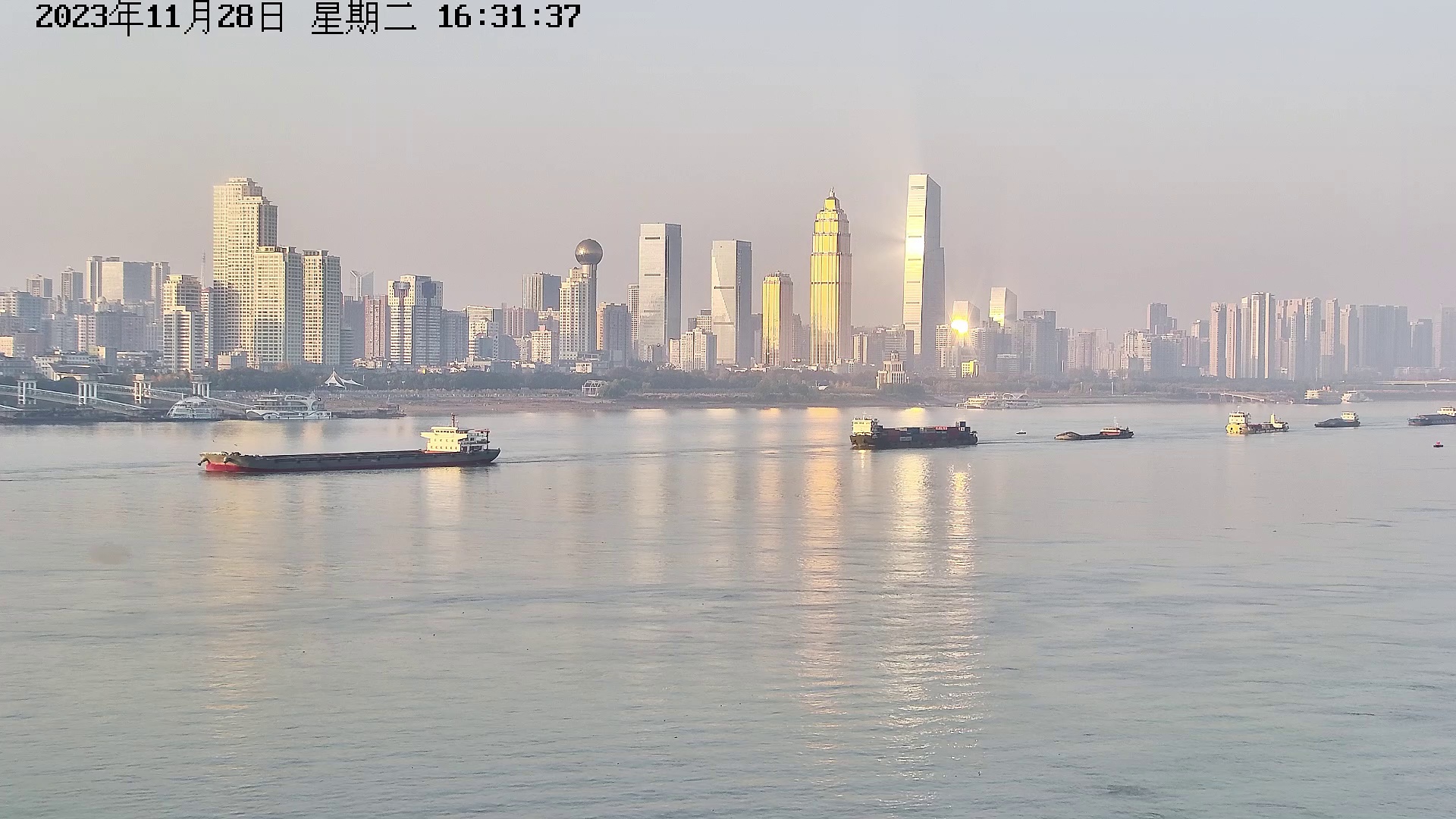}
        \label{fig:multi_light_2}
    \end{subfigure}
    \hfill
    \begin{subfigure}{0.24\textwidth}
        \includegraphics[width=\textwidth]{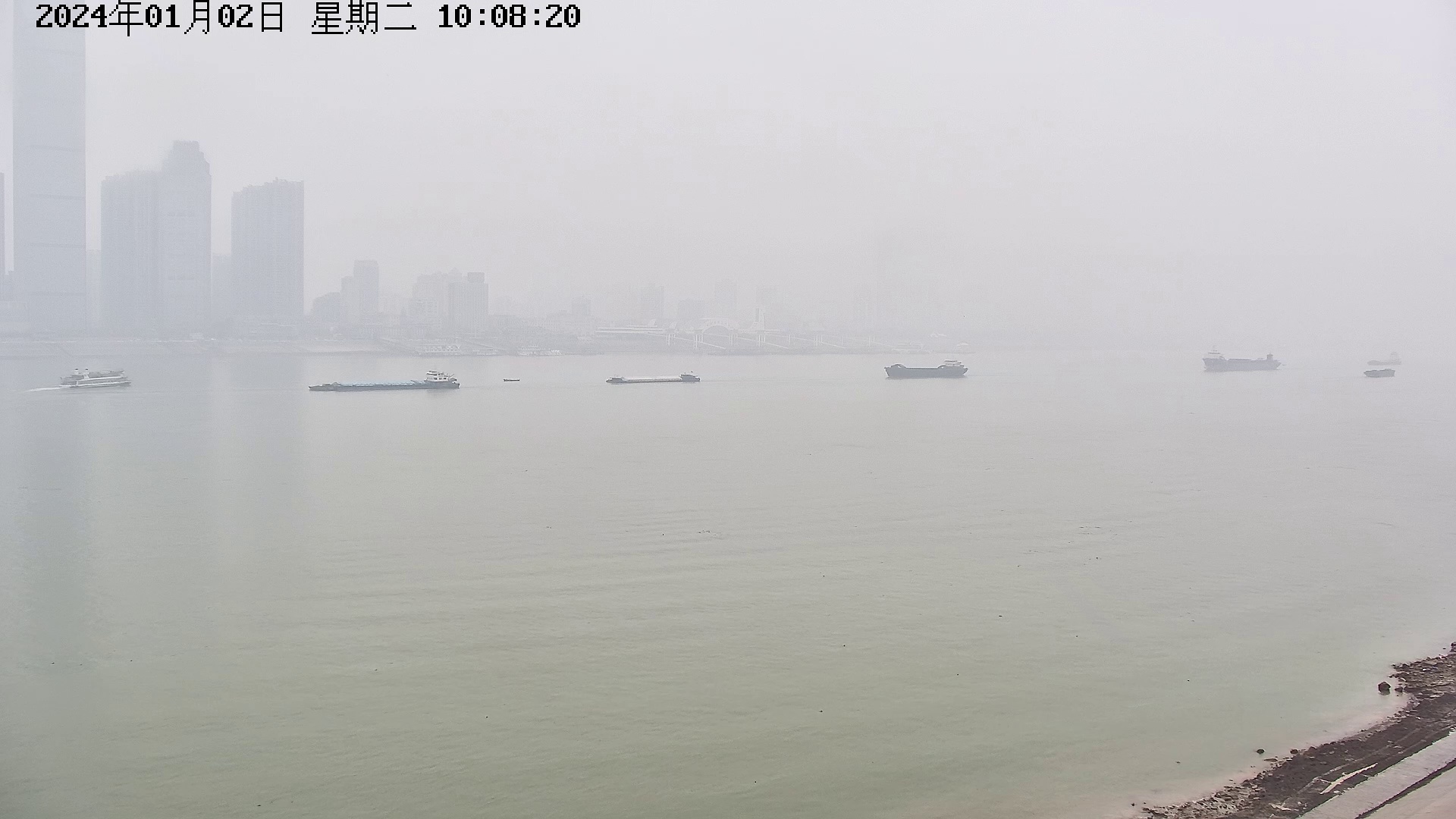}
        \label{fig:multi_weather_2}
    \end{subfigure}
    \hfill
    \begin{subfigure}{0.24\textwidth}
        \includegraphics[width=\textwidth]{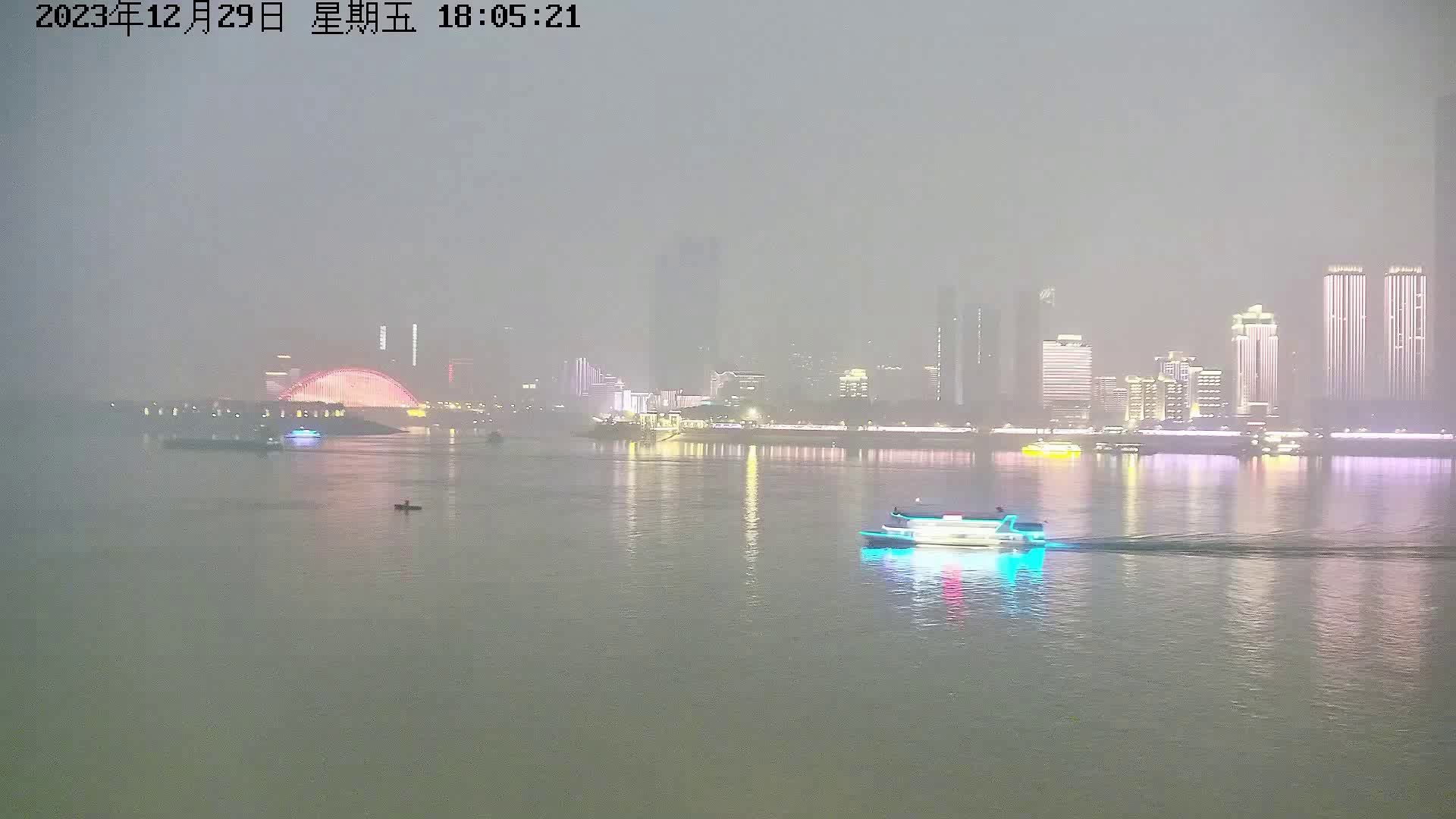}
        \label{fig:multi_scenario_2}
    \end{subfigure}
    \\ % 换行
    \vspace{-1em}
    % 第三行
    \begin{subfigure}{0.24\textwidth}
        \includegraphics[width=\textwidth]{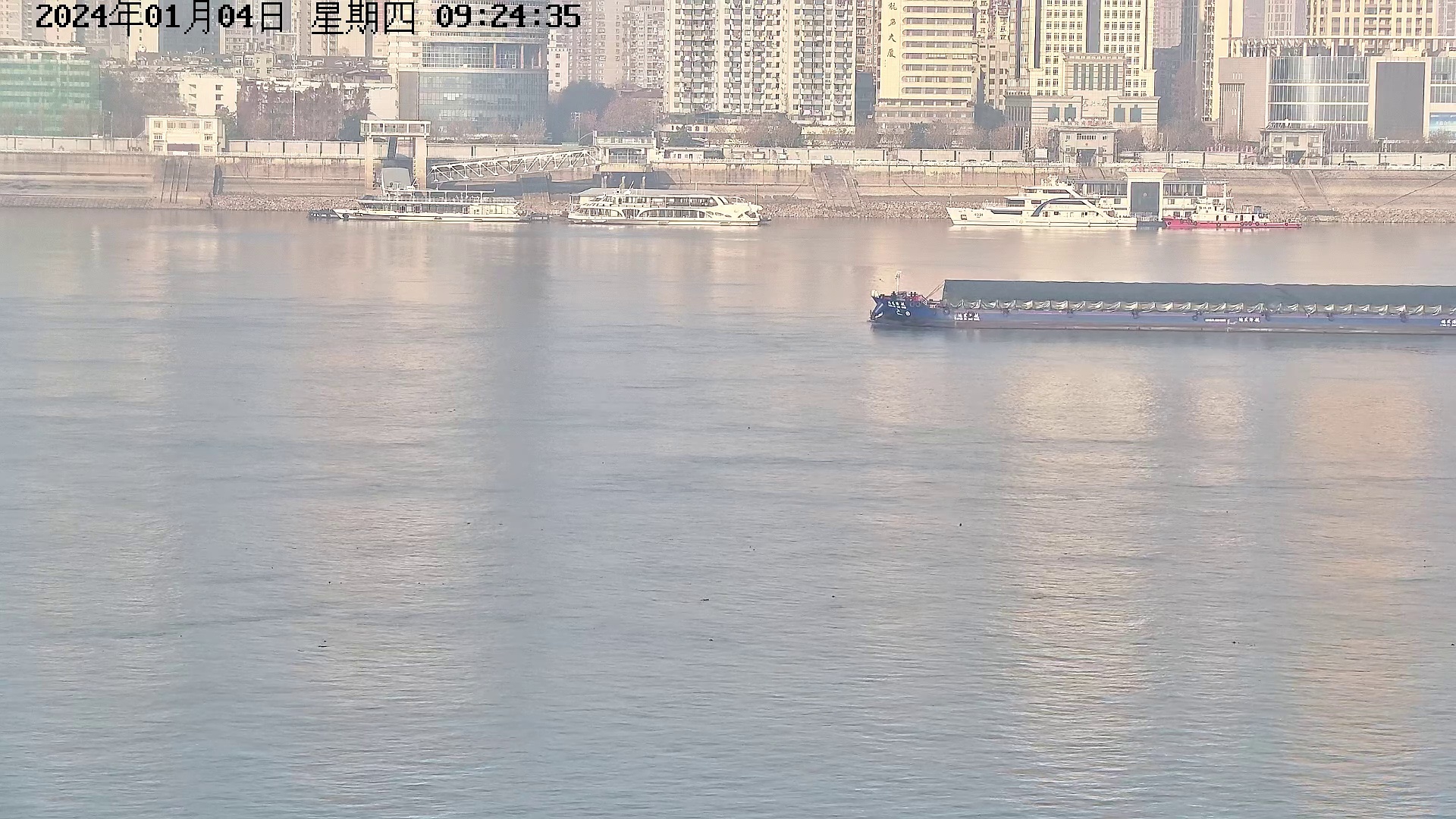}
        \label{fig:multi_view_3}
    \end{subfigure}
    \hfill
    \begin{subfigure}{0.24\textwidth}
        \includegraphics[width=\textwidth]{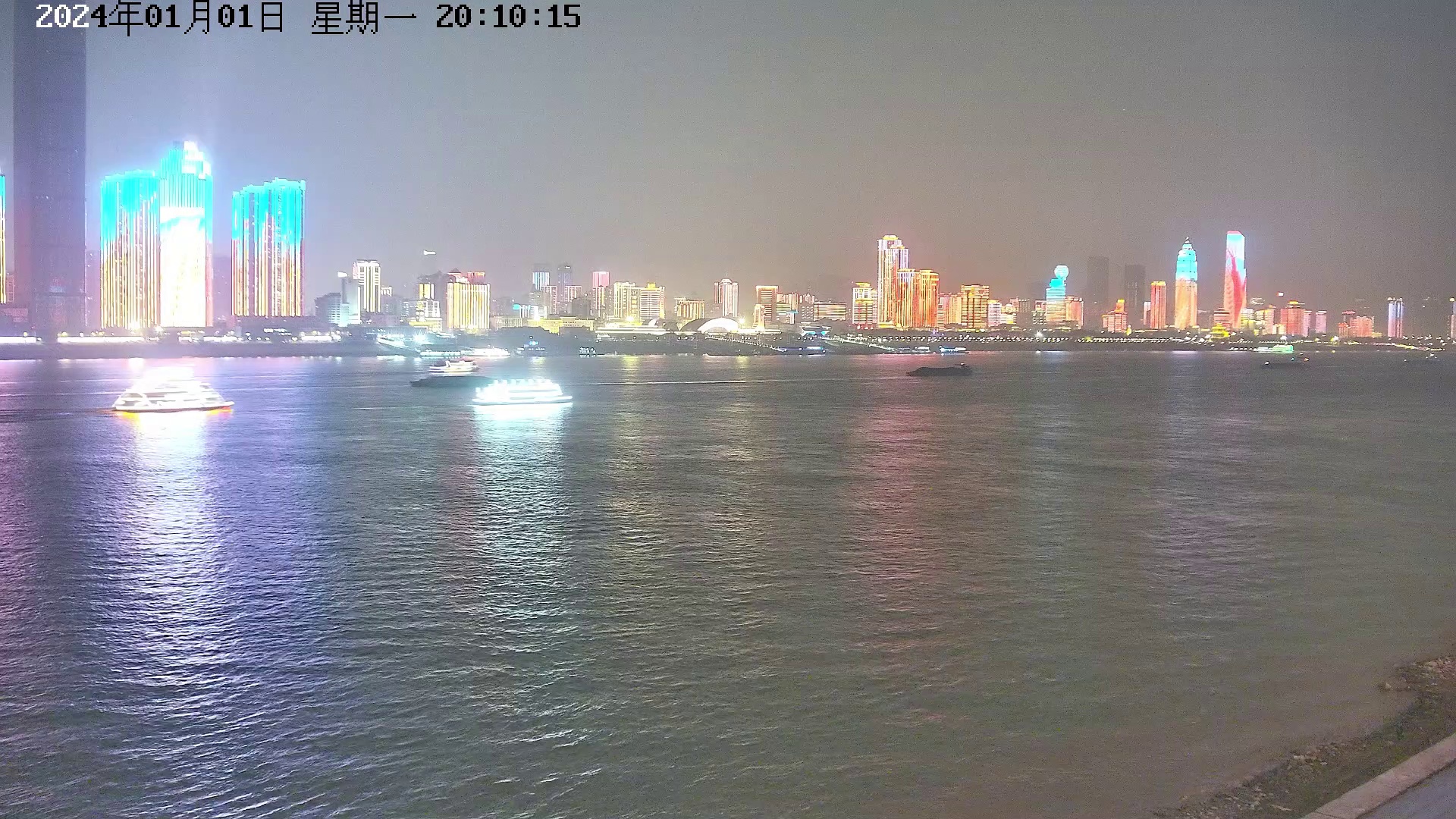}
        \label{fig:multi_light_3}
    \end{subfigure}
    \hfill
    \begin{subfigure}{0.24\textwidth}
        \includegraphics[width=\textwidth]{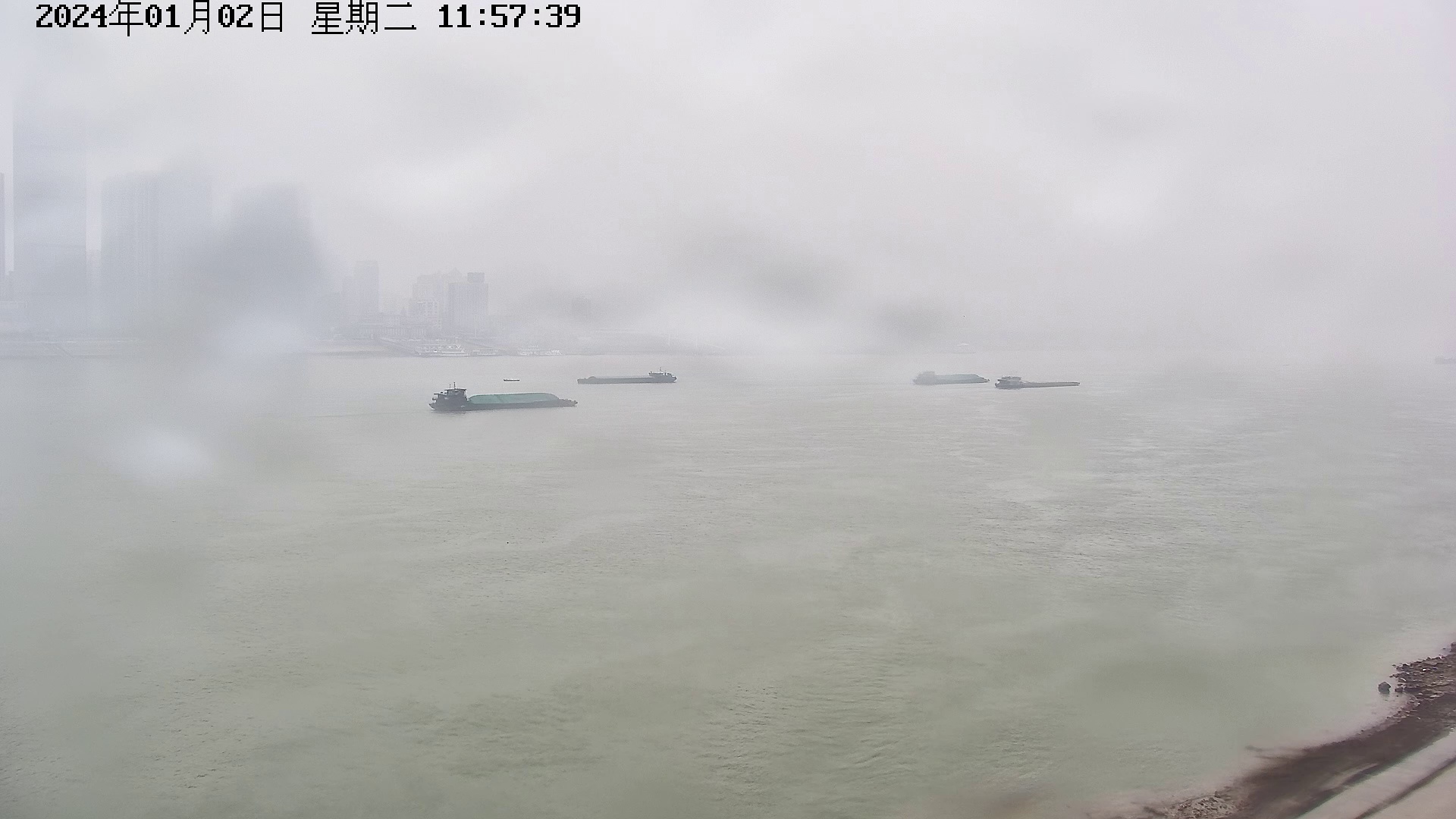}
        \label{fig:multi_weather_3}
    \end{subfigure}
    \hfill
    \begin{subfigure}{0.24\textwidth}
        \includegraphics[width=\textwidth]{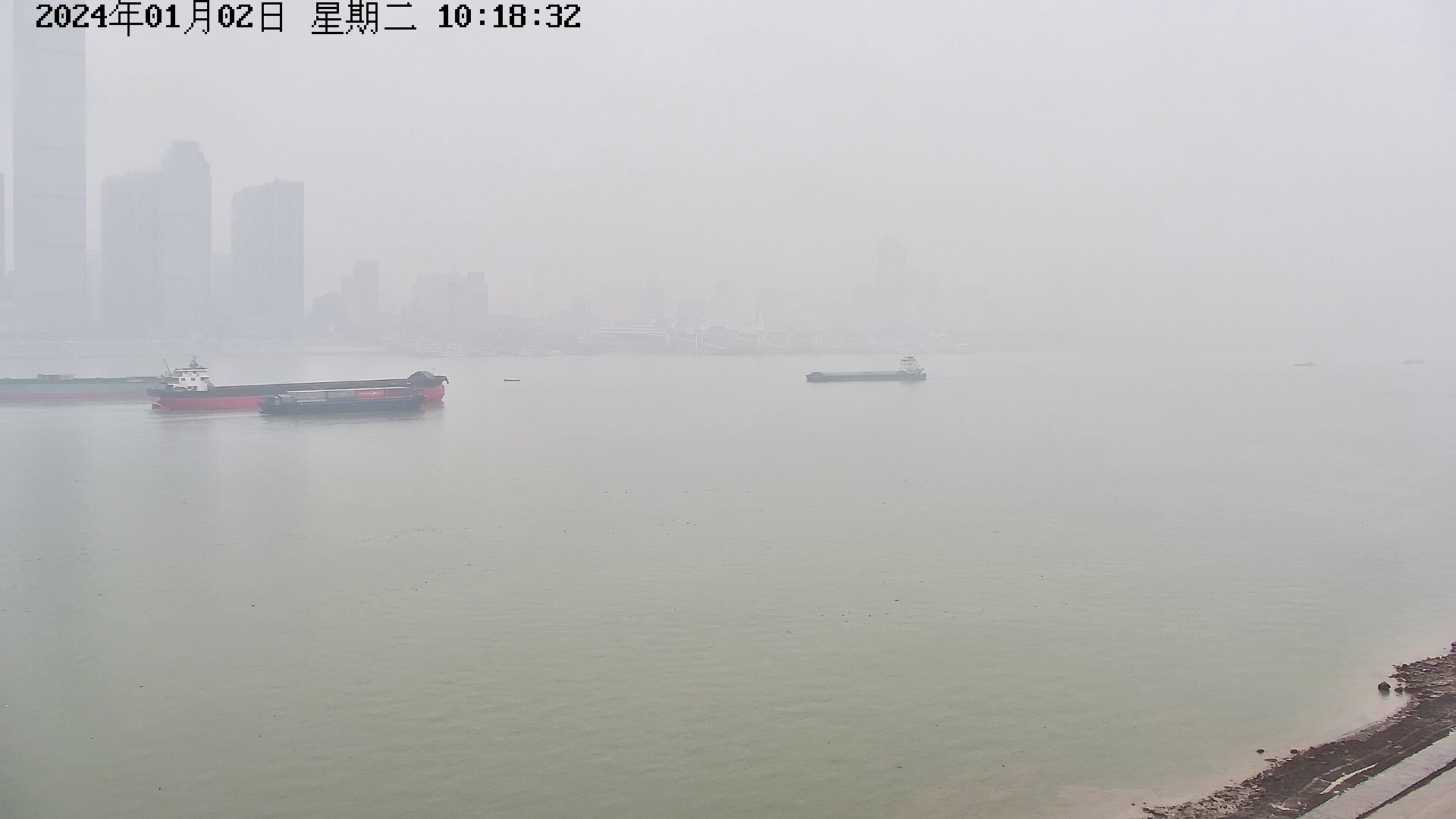}
        \label{fig:multi_scenario_3}
    \end{subfigure}
    \begin{minipage}{\textwidth}
        \centering
        (a) Multi-view \quad (b) Multi-light \quad (c) Multi-weather \quad (d) Multi-scenario
    \end{minipage}
    % 小标题
    \caption{Representative samples illustrating diverse viewing angles, illumination conditions, meteorological variations, and scenarios from MEIWVD.}
    \label{fig:representative_samples}
    \par\medskip
\end{figure}

The MEIWVD spans a total of six months from November 2023 to April 2024, including video clips from multiple perspectives, lighting and weather conditions. Through preprocessing, we carefully selected 119 clips rich in surface objects and annotated the data using the DarkLabel tool. Example images illustrating these features are shown in \ref{fig:representative_samples}. The specific features include:
\begin{enumerate}[label=(\arabic*)]
	\item Multiple perspectives: The dataset covers surface objects from front, rear, and side views to increase data diversity. 
	\item Multiple lighting conditions: Candidate images were captured under various lighting conditions, such as strong light, low light, and artificial lighting, to reflect the impact of different lighting conditions on the natural environment. 
	\item Multiple weather conditions: The dataset encompasses diverse meteorological conditions, including sunny, cloudy, rainy, foggy, etc., to ensure comprehensive environmental representation.
	\item Multiple scenarios: The diversity of occlusions and backgrounds was considered.
\end{enumerate}
\subsection{Dataset analysis}
\label{sec:sec3.2}
\subsubsection{Data collection strategy based on synergy of natural and urban lighting}
\label{sec:sec3.2.1}
\begin{figure}[htbp]
    \centering
    \includegraphics[width=0.8\textwidth]{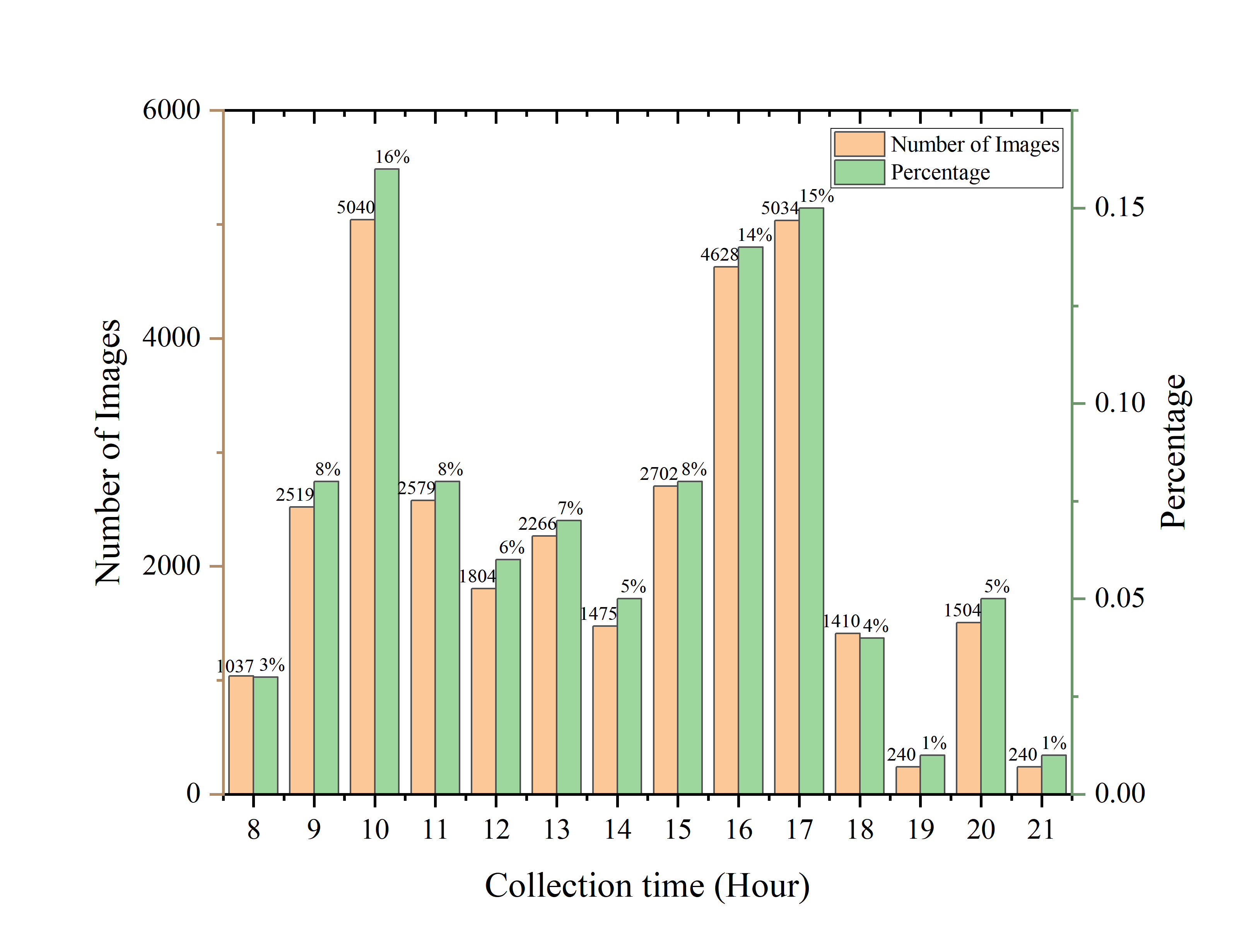} % 加载图片，宽度为页面宽度的 80%
    \caption{Temporal distribution and percentage representation of image acquisition time points in the MEIWVD.} % 图片标题
    \label{fig:collection_time} % 图片标签
\end{figure}

To accurately reflect the real-world scenarios of the Yangtze River inland waterways, we conducted long-term data collection of surface objects. The collection period was set from 8:00 AM to 6:00 PM daily to ensure the dataset realistically represents actual lighting conditions. Additionally, considering that the inland river areas are often accompanied by urban light shows and lights from passenger ships, these artificial light sources can effectively supplement the lack of natural light during the evening and nighttime. Therefore, the dataset intentionally includes data collected from 6:00 PM to 9:00 PM to extend the operational range of visible light. \ref{fig:collection_time} illustrates the time periods covered by the collected data. As shown in \ref{fig:collection_time}, the data collection in this dataset spans from 8:00 AM to 9:00 PM, with the highest data volume observed at 10:00 AM and between 4:00 PM and 5:00 PM.

\subsubsection{Multi-environment analysis}
\label{sec:sec3.2.2}
\begin{figure}[htbp]
    \centering
    \includegraphics[width=0.8\textwidth]{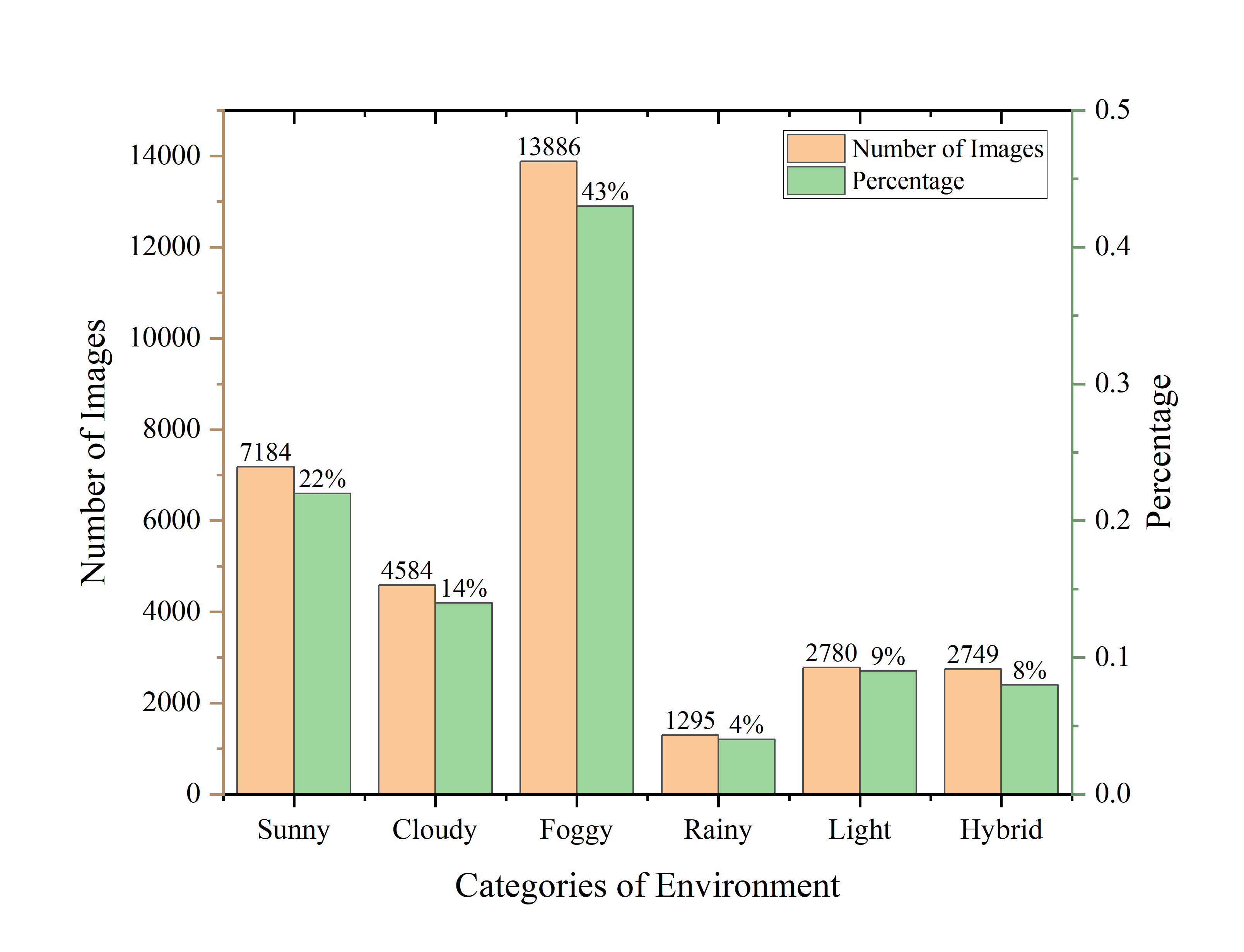} % 加载图片，宽度为页面宽度的 80%
    \caption{Temporal distribution and percentage representation of image acquisition time points in the MEIWVD.} % 图片标题
    \label{fig:multi_environment} % 图片标签
\end{figure}

\begin{figure}[htbp]
    \centering
    \includegraphics[width=\textwidth]{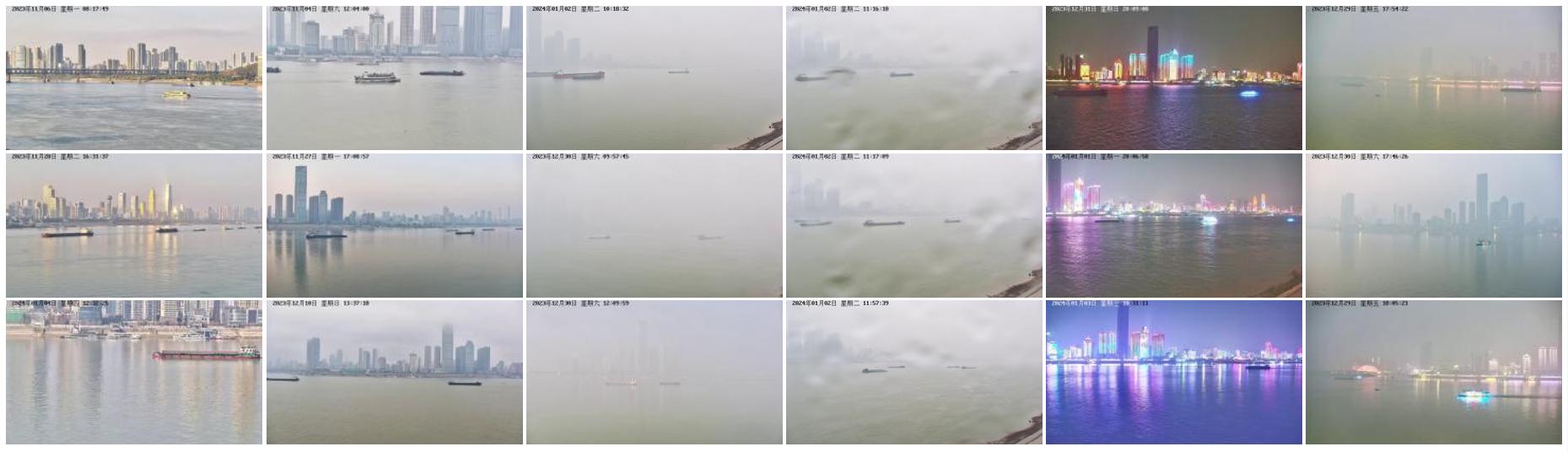} % 加载图片，宽度为页面宽度的 80%
    \vspace{-0.5em} % 调整标题与标注之间的间距
    \begin{center}
        \small
        (a) Sunny \quad (b) Cloudy \quad (c) Foggy \quad (d) Rainy \quad (e)Light \quad (f) Hybrid
    \end{center}
    \caption{Examples of multi-environment scenarios in the MEIWVD.} % 图片标题
    \label{fig:example_environment} % 图片标签
\end{figure}

After meticulous screening, we selected real-world data encompassing various weather conditions, including natural lighting, clear skies, fog, and rain. \ref{fig:multi_environment} illustrates the distribution of collected images across different categories. Images captured under clear weather conditions amount to 7,184, constituting 22.1\% of the dataset. Foggy conditions, due to their frequent occurrence in inland waterways, comprise 13,886 images, representing 42.1\%. Overcast conditions account for 4,584 images, corresponding to 14.1\%, while rainy conditions consist of 1,295 images, representing 4.0\%. Images featuring urban light shows and passenger ships light total 2,780, occupying 8.6\% of the dataset. Additionally, images that include both lighting and fog conditions amount to 2,749, accounting for 8.5\%. Notably, foggy images dominate the dataset, reflecting the real-world prevalence of fog in inland waterways. Example images of multi-environment conditions are illustrated in \ref{fig:example_environment}.  

\subsubsection{Category analysis of  surface objects}
\label{sec:sec3.2.3}
The collected data were analyzed to determine the specific categories of surface objects included in the MEIWVD, along with the corresponding names and sample counts for each category, as shown in Table \ref{tab:num_surface_objects}. In total, MEIWVD contains 32,478 precisely annotated images. Among these, cargo ships, which are the most common in the Yangtze River basin, account for 87,303 objects, representing 45.9\% of the dataset. Passenger ships follow, with 68,192 objects, making up 35.9\%. Buoys, which are also prevalent, total 31,814 objects, constituting 16.7\%. The least common category is container ships, with only 2,883 objects, representing 1.5\%. The distribution of surface object categories exhibits a reasonable imbalance. In inland waterways, buoys are a common type of surface object. Although smaller in size compared to ships, they are critically important in object detection. Buoys and cargo ships represent objects with extreme size differences in the MEIWVD. 

\begin{table}[htbp]
    \centering
    \caption{Number of surface objects in each category.}
    \label{tab:surface_objects}
    \begin{tabular}{lrr}
    \hline
    \rule{0pt}{1.5em}\textbf{Category} & \textbf{Number of objects} & \textbf{Percentage} \\ \hline
    Cargo ship        & 87,303                     & 45.9\%             \\ 
    Passenger ship    & 68,192                     & 35.9\%             \\ 
    Buoy              & 31,814                     & 16.7\%             \\ 
    Container ship    & 2,883                      & 1.5\%              \\ \hline
    \textbf{Total}             & 190,192                    & 100\%              \\ \hline
    \end{tabular}
\label{tab:num_surface_objects}%
\end{table}

\begin{table}[htbp]
	\centering
	\caption{Absolute multi-scale distribution of surface objects in the SeaShips and MEIWVD.}
	\begin{tabular}{cp{10.61em}p{6em}c}
		\toprule
		\multirow{2}[4]{*}{} & \multicolumn{1}{c}{\multirow{2}[4]{*}{Absolute scale}} & \multicolumn{2}{c}{Statistics} \\
		\cmidrule{3-4}          & \multicolumn{1}{c}{} & \multicolumn{1}{p{9em}}{Number of objects} & \multicolumn{1}{p{4.055em}}{Percentage} \\
		\midrule
		\multicolumn{1}{c}{\multirow{4}[2]{*}{SeaShips}} & Small & 35 & 0.37\% \\
		& Medium & 934 & 9.94\% \\
		& Large & \textbf{8,429} & \textbf{89.69\%} \\
		\cmidrule{2-4}
		& Total & 9,398 & 100\% \\
		\midrule
		\multicolumn{1}{c}{\multirow{4}[2]{*}{MEIWVD}} & Small & \textbf{55,027} & \textbf{28.93\%} \\
		& Medium & \textbf{111,093} & \textbf{58.41\%} \\
		& Large & 24,072 & 12.66\% \\
		\cmidrule{2-4}
		& Total & \textbf{190,192} & 100\% \\
		\bottomrule
	\end{tabular}%
\label{tab:absolute_multi_scale_distribution}% multi_scale_distribution
\end{table}%

\subsubsection{Multi-scale analysis of surface objects}
\label{sec:sec3.2.4}
In the field of object detection, multi-scale detection refers to the capability of algorithms to identify and process objects of varying sizes. A significant challenge in this area is the considerable scale variation among surface objects, particularly the detection of small objects. Small objects can be defined in two ways by absolute scale or relative scale. In the MS COCO dataset \cite{linMicrosoftCOCOCommon2014}, small objects are those with a resolution of less than 32\texttimes32 pixels. Alternatively, some researchers define small objects based on the ratio of the bounding box dimensions to the image dimensions, specifically when this ratio is less than a certain threshold (e.g., 0.1), or when the square root of the ratio of the bounding box area to the image area is below a specified value.

To analyze the multi-scale characteristics of the datasets, we examined the distribution differences of multi-scale objects in the SeaShips and MEIWVD. Given that surface objects often exhibit elongated shapes, Table \ref{tab:absolute_multi_scale_distribution} presents the proportions of objects at different scales in both datasets. Small objects are defined as having a resolution of less than 32\texttimes32 pixels, medium objects as less than 96\texttimes96 pixels, and large objects as greater than 96\texttimes96 pixels. in the SeaShips, small-scale objects account for 0.37\%, while in the MEIWVD, they comprise 28.93\%. Conversely, large-scale objects constitute 89.69\% in the SeaShips but only 12.66\% in the MEIWVD. From an absolute scale perspective, objects in the MEIWVD are generally smaller, which aligns more closely with real-world engineering applications. Furthermore, MEIWVD contains a greater number of images and a higher total sample count compared to SeaShips.

% \begin{table}[htbp]
% 	\centering
% 	\caption{Absolute multi-scale distribution of surface objects in the SeaShips and MEIWVD.}
% 	\begin{tabular}{lp{10.61em}p{6em}c}
% 		\toprule
% 		\multirow{2}[4]{*}{} & \multicolumn{1}{c}{\multirow{2}[4]{*}{Absolute scale}} & \multicolumn{2}{c}{Statistics} \\
% 		\cmidrule{3-4}          & \multicolumn{1}{c}{} & \multicolumn{1}{p{10em}}{Number of objects} & \multicolumn{1}{p{4.055em}}{Percentage} \\
% 		\midrule
% 		\multicolumn{1}{l}{\multirow{4}[2]{*}{SeaShips}} & Small & 35 & 0.37\% \\
% 		& Medium & 934 & 9.94\% \\
% 		& Large & \textbf{8,429} & \textbf{89.69\%} \\
% 		& Total & 9,398 & 100\% \\
% 		\midrule
% 		\multicolumn{1}{l}{\multirow{4}[2]{*}{MEIWVD}} & Small & \textbf{55,027} & \textbf{28.93\%} \\
% 		& Medium & \textbf{111,093} & \textbf{58.41\%} \\
% 		& Large & 24,072 & 12.66\% \\
% 		& Total & 190,192 & 100\% \\
% 		\bottomrule
% 	\end{tabular}%
% \label{tab:absolute_multi_scale_distribution}%
% \end{table}%

\begin{figure}[htbp]
    \centering
    % 第一行
    \subcaptionbox{{\fontsize{8}{8}\selectfont Width-ratio in the SeaShips}\label{fig:width_seaships}}[0.3\textwidth]{
        \includegraphics[width=\linewidth]{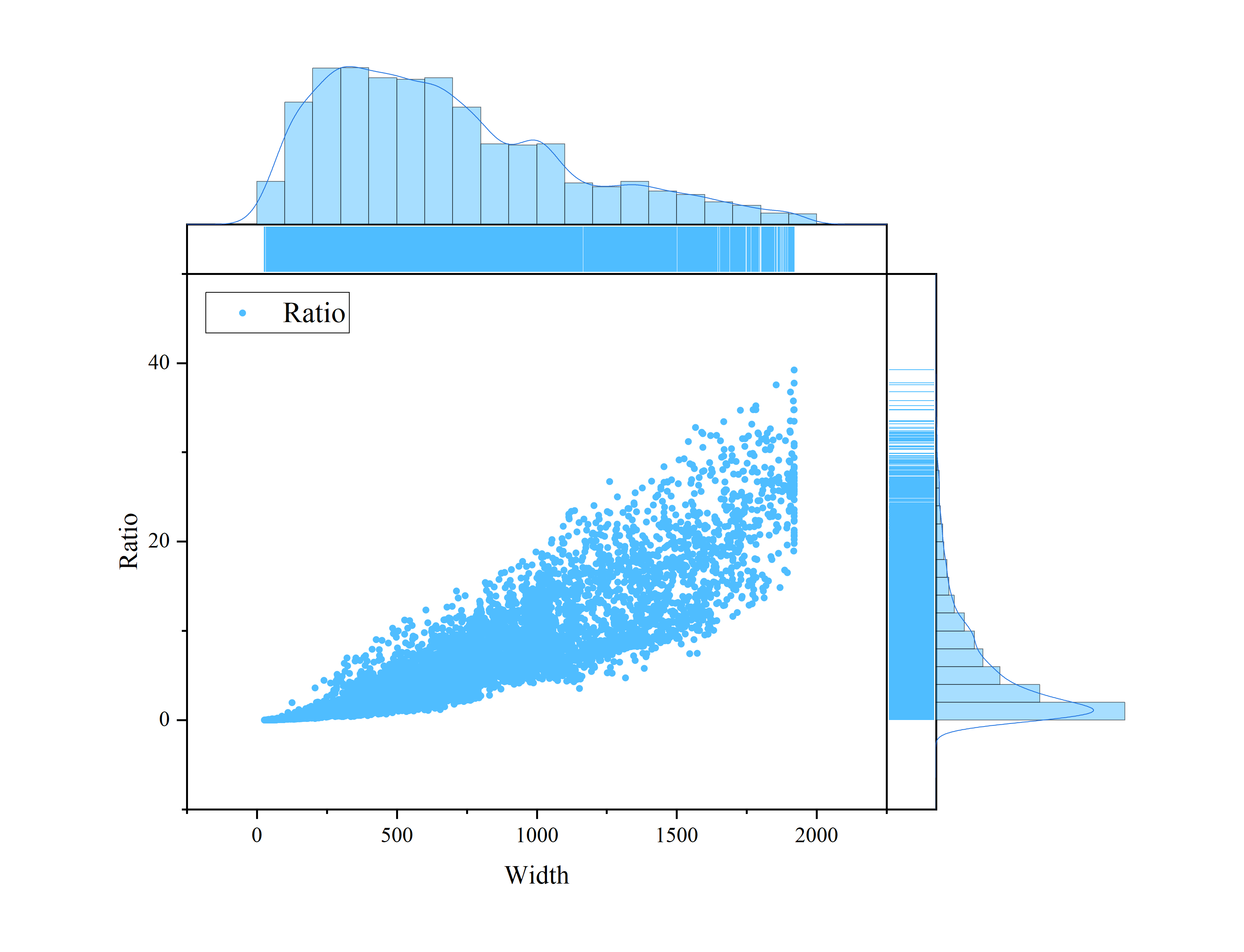}
    }
    \hfill
    \subcaptionbox{{\fontsize{8}{8}\selectfont Height-ratio in the SeaShips}\label{fig:height_seaships}}[0.3\textwidth]{
        \includegraphics[width=\linewidth]{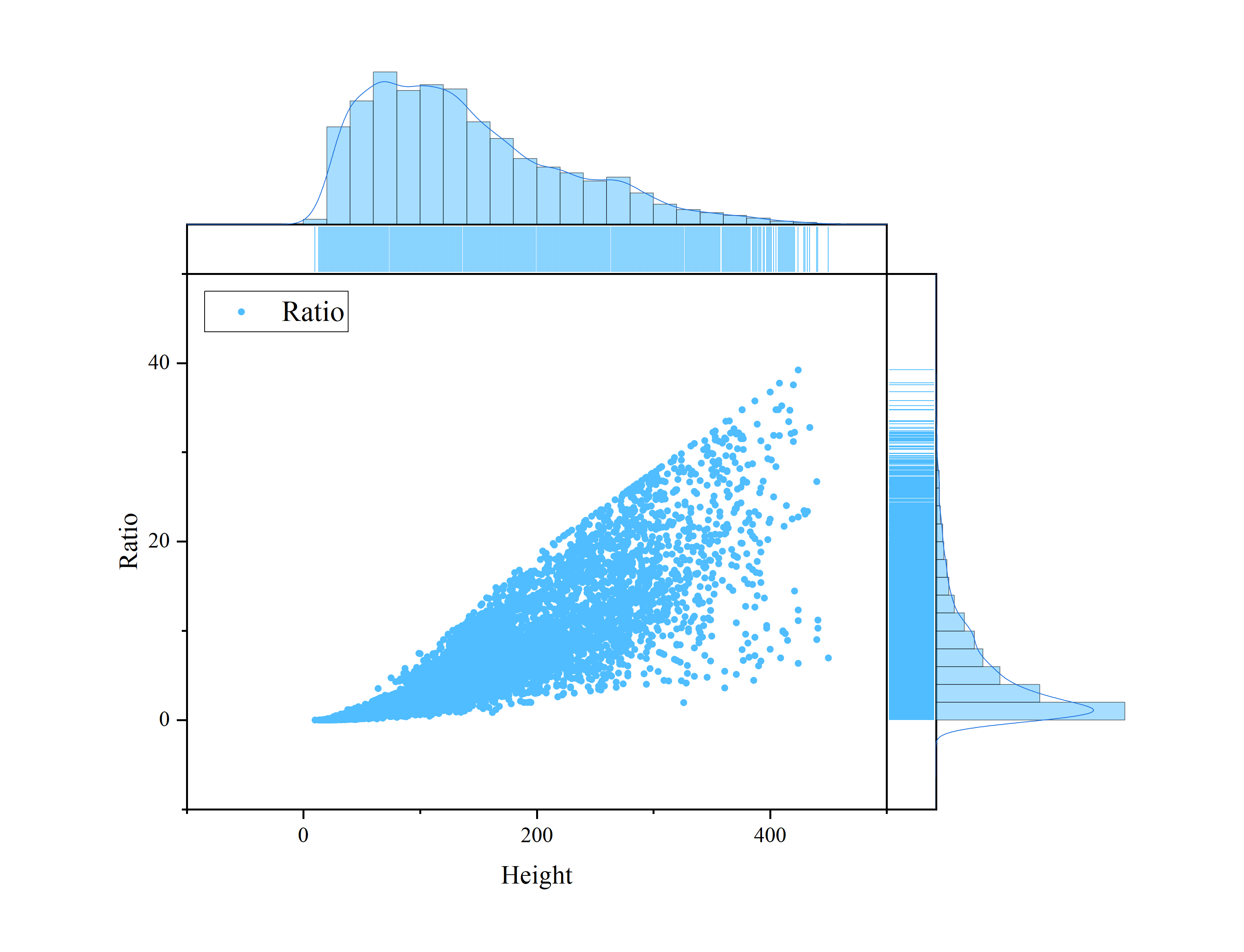}
    }
    \hfill
    \subcaptionbox{{\fontsize{8}{8}\selectfont Normal probability plot of ratio in the SeaShips}\label{fig:prob_seaships}}[0.33\textwidth]{
        \includegraphics[width=\linewidth]{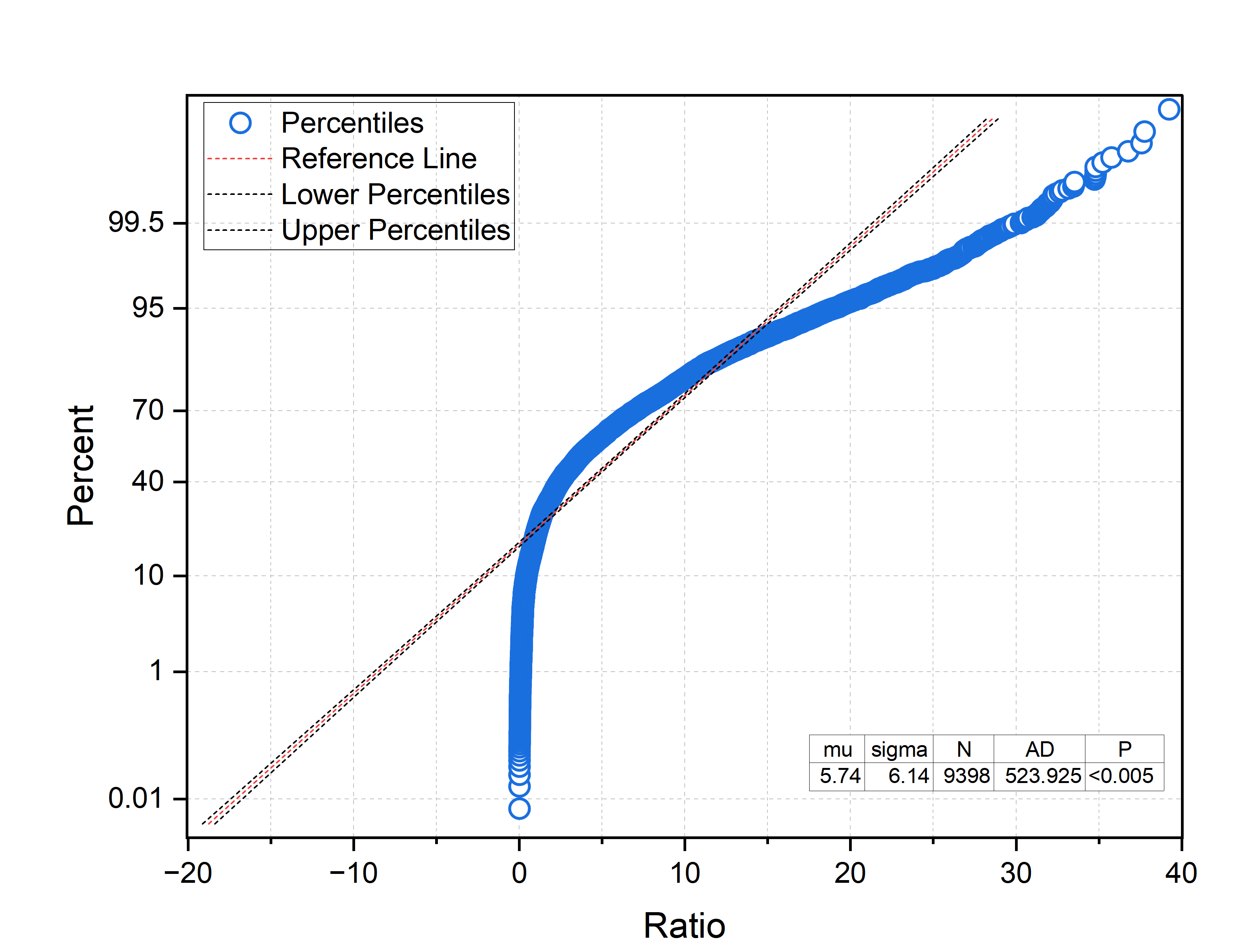}
    }
    % 第二行
    \subcaptionbox{{\fontsize{7}{8}\selectfont Width-ratio in the MEIWVD}\label{fig:width_meiwvd}}[0.3\textwidth]{
        \includegraphics[width=\linewidth]{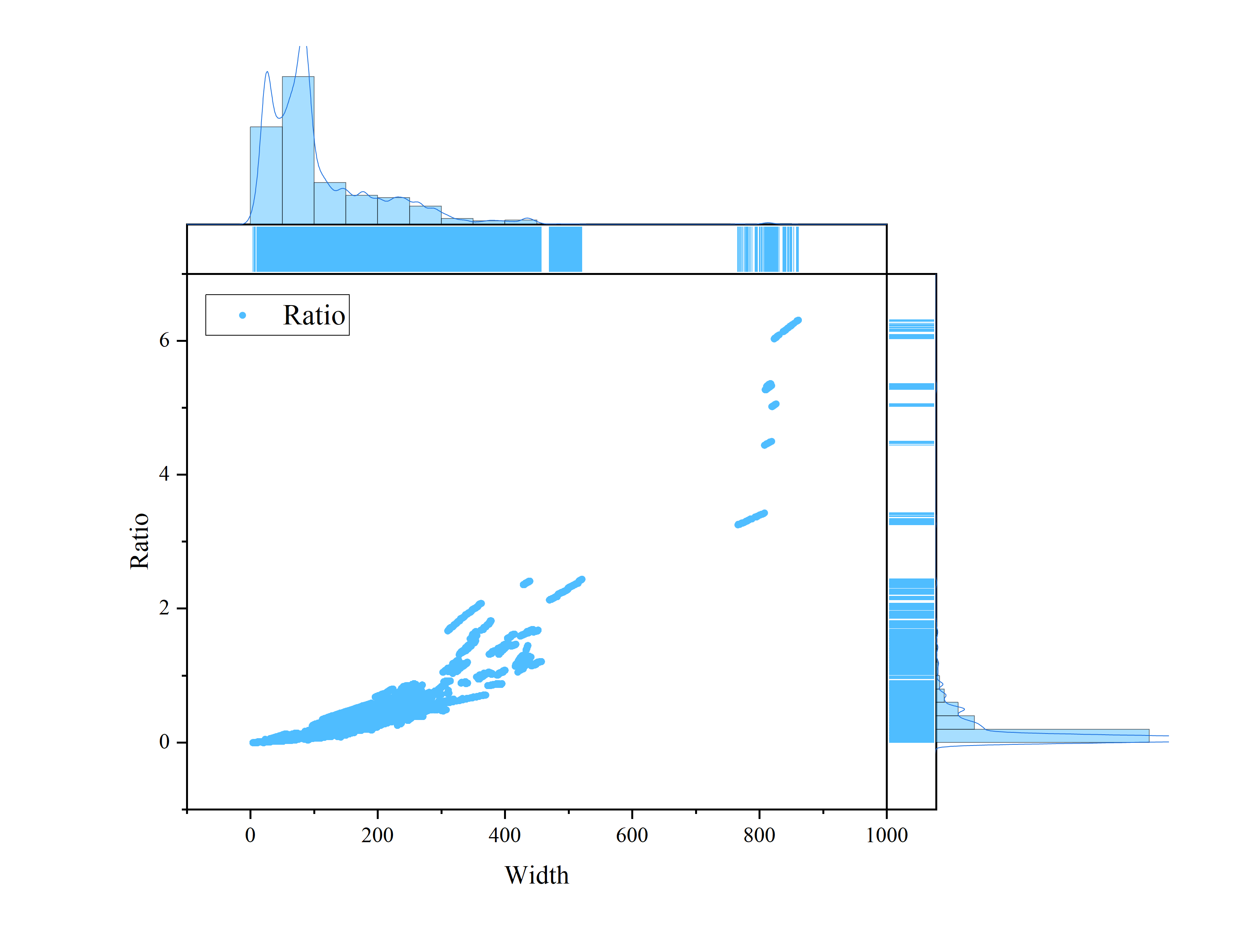}
    }
    \hfill
    \subcaptionbox{{\fontsize{7}{8}\selectfont Height-ratio in the MEIWVD}\label{fig:height_meiwvd}}[0.3\textwidth]{
        \includegraphics[width=\linewidth]{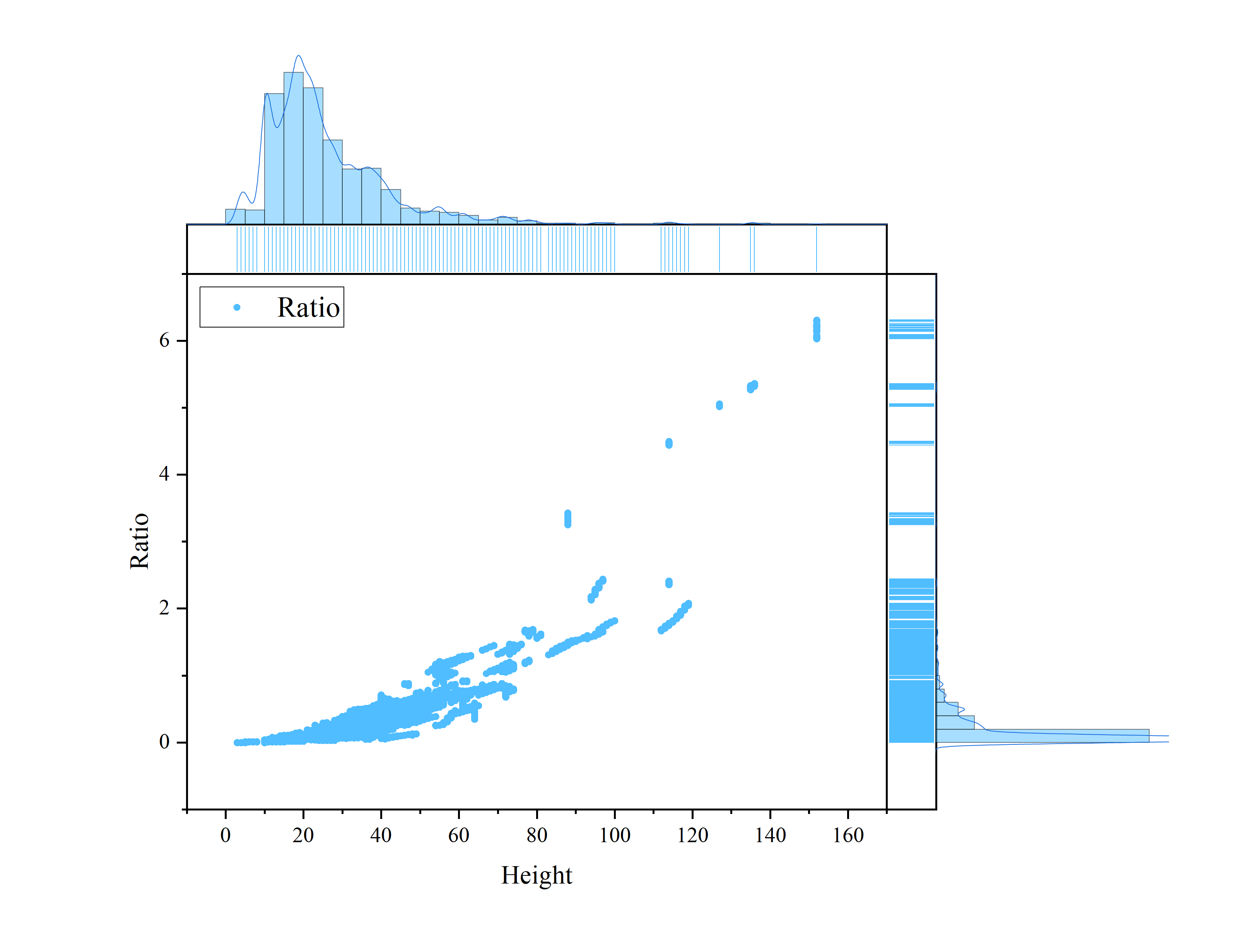}
    }
    \hfill
    \subcaptionbox{{\fontsize{8}{8}\selectfont Normal probability plot of ratio in the MEIWVD}\label{fig:prob_meiwvd}}[0.33\textwidth]{
        \includegraphics[width=\linewidth]{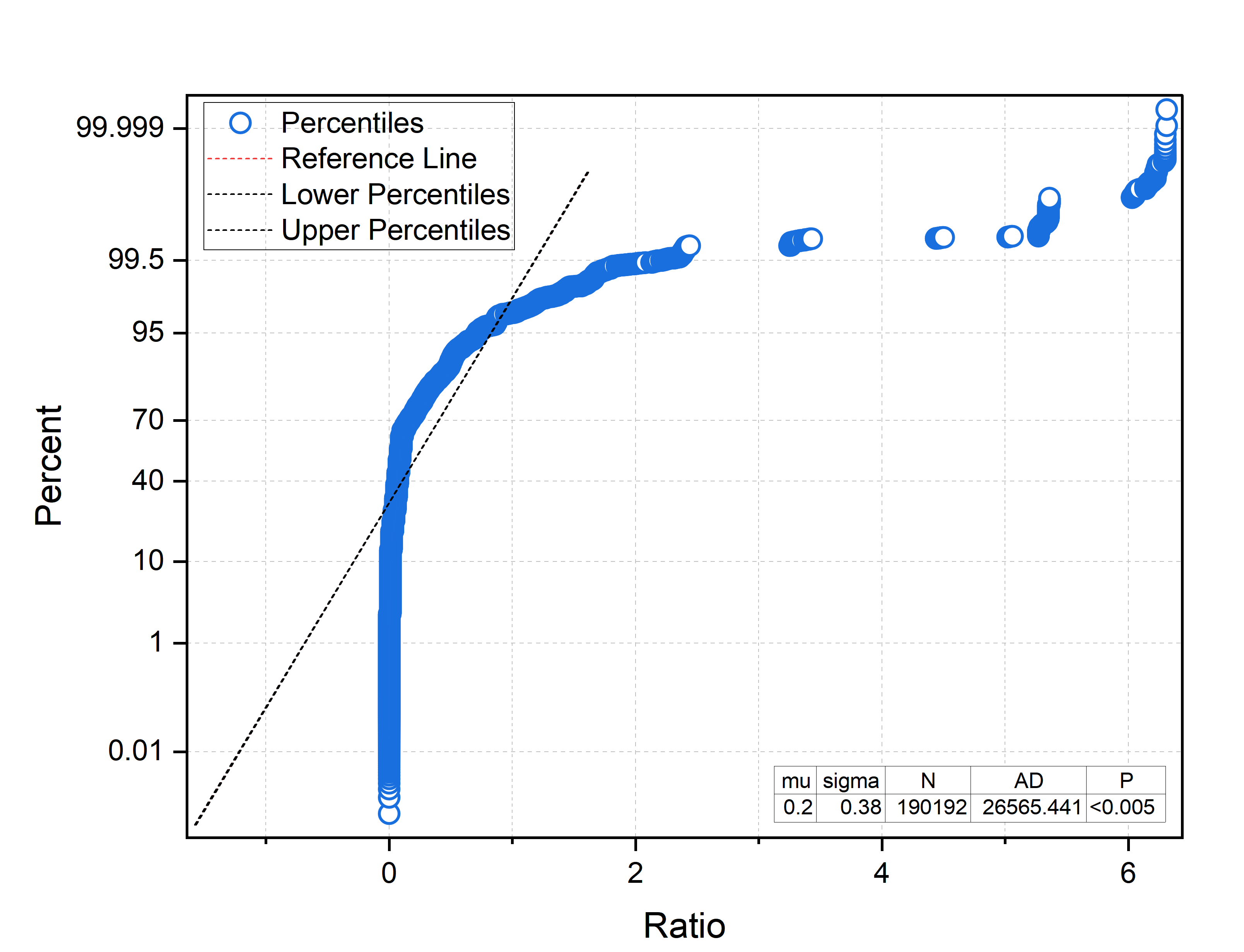}
    }
    % 总标题
    \caption{Relative multi-scale distribution of surface objects in the SeaShips and MEIWVD.} % 总标题
    \label{fig:multi_scale_distribution}
\end{figure}

\ref{fig:multi_scale_distribution} illustrates the relative scale distribution of surface objects in the SeaShips and MEIWVD. The vertical axis, labeled Ratio represents the proportion of the bounding box area of each surface object relative to the total area of the image. Specifically, \ref{fig:width_seaships} and \ref{fig:height_seaships} present scatter plots and histograms of the width and height of surface objects in  SeaShips against the ratio, while \ref{fig:width_meiwvd} and \ref{fig:height_meiwvd} depict the corresponding plots for MEIWVD. \ref{fig:prob_seaships} and \ref{fig:prob_meiwvd} exhibit the probability distributions of the ratio for both datasets.

From \ref{fig:width_seaships}-\ref{fig:height_seaships} and \ref{fig:width_meiwvd}-\ref{fig:height_meiwvd}, we can see that surface objects in the SeaShips are generally larger than those in the MEIWVD. \ref{fig:prob_seaships} shows that 99.5\% of objects have a relative scale of less than 2\% in the MEIWVD, whereas this percentage reaches 99.5\% for objects with a relative scale of up to 35\% in the SeaShips. Therefore, the detection of smaller surface objects in the MEIWVD presents unique challenges that require tailored approaches for effective identification.

\begin{figure}[htbp]
    \centering
    % 第一列
    \begin{subfigure}{0.48\textwidth}
        \centering
        \includegraphics[width=\linewidth]{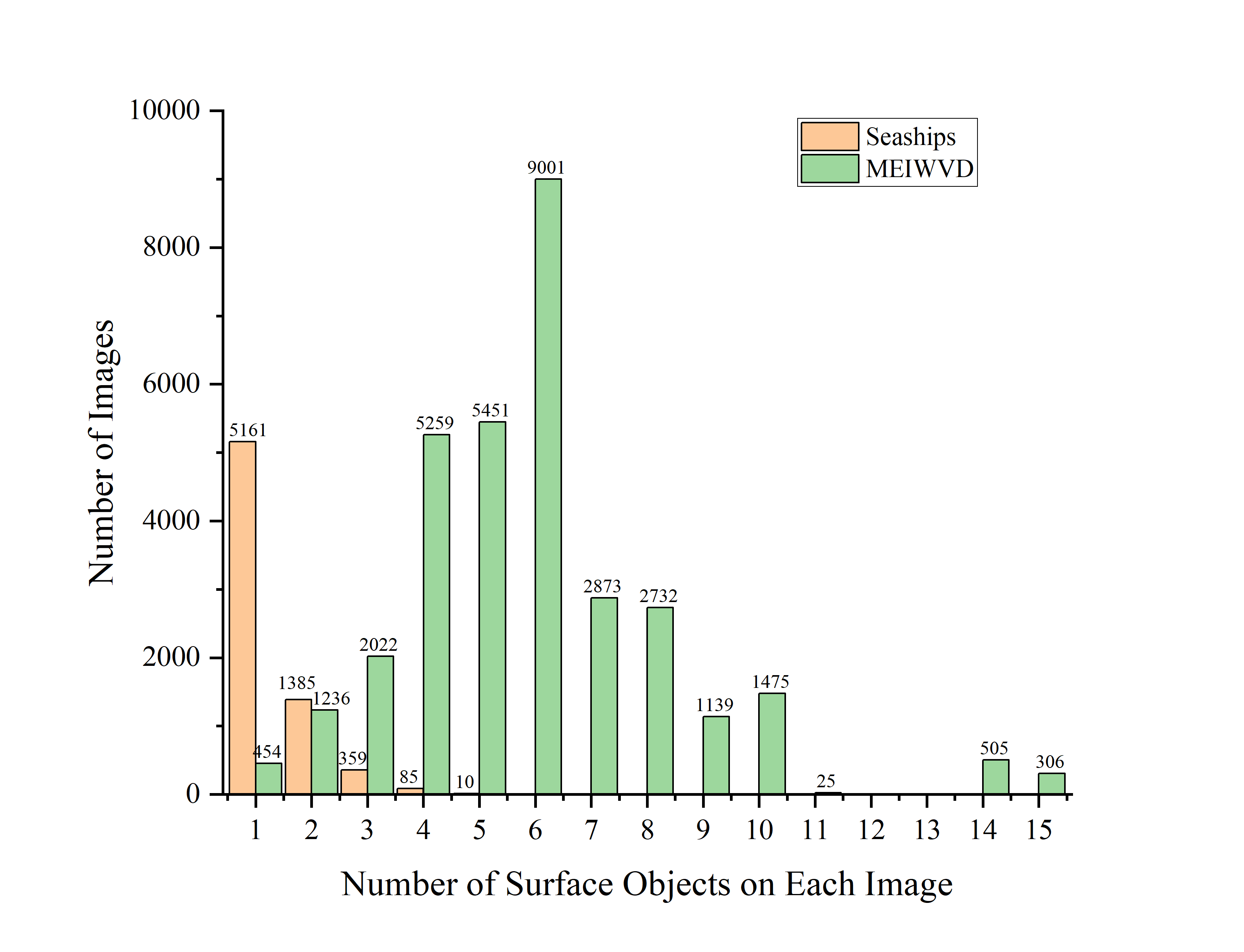} % 使用示例图片
        \caption{Distribution of surface objects per image in the datasets.} % 子图标题
        \label{fig:dist_per_image}
    \end{subfigure}
    \hfill
    % 第二列
    \begin{subfigure}{0.48\textwidth}
        \centering
        \includegraphics[width=\linewidth]{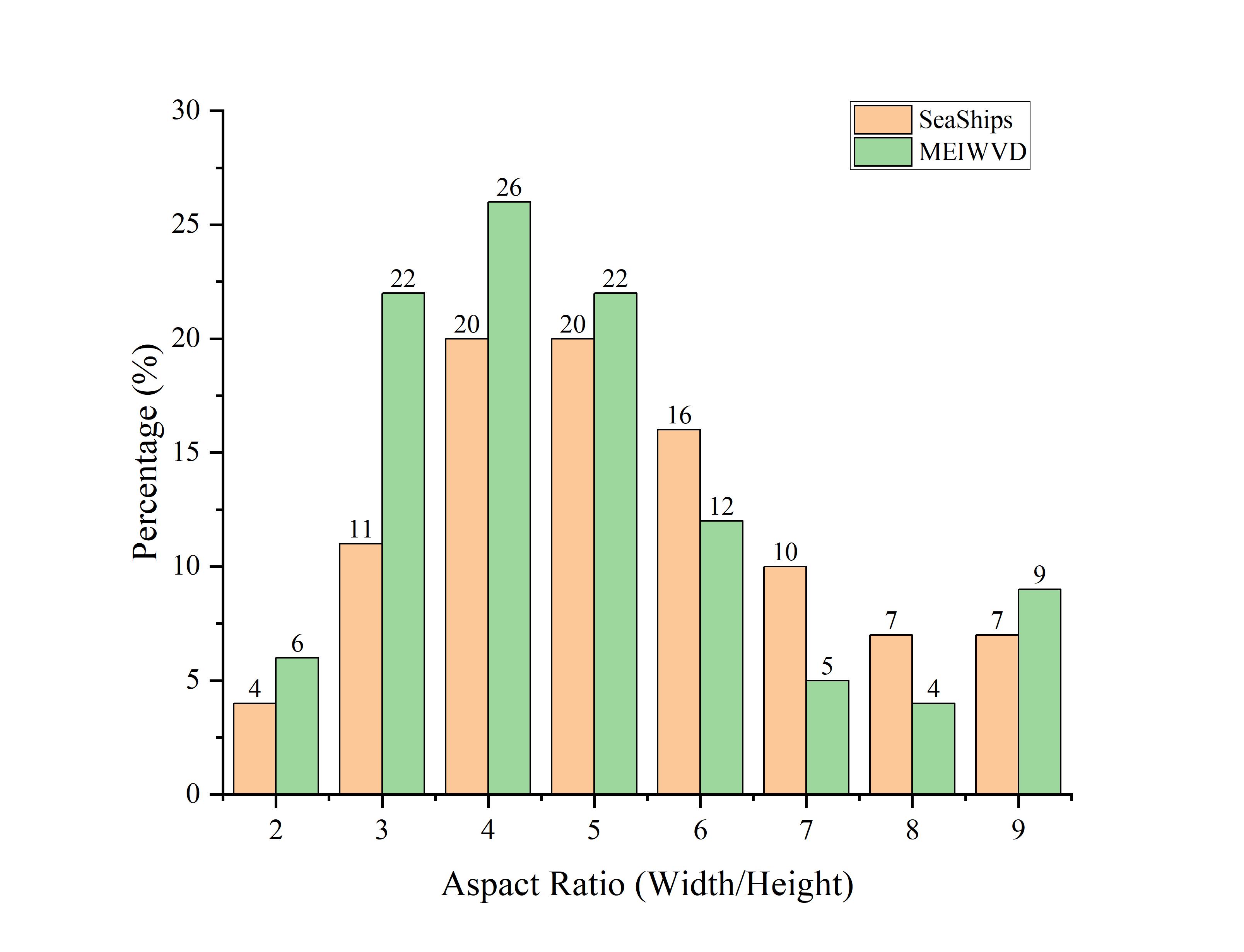} % 使用示例图片
        \caption{Percentage of aspect ratios of surface objects in the datasets.} % 子图标题
        \label{fig:aspect_ratio}
    \end{subfigure}
    % 总标题
    \caption{Distribution of surface objects in the datasets.} % 总标题
    \label{fig:dist_surface_objects}
\end{figure}

We further analyze the number of surface objects per image in the datasets, as illustrated in \ref{fig:dist_surface_objects}. In \ref{fig:dist_per_image}, the horizontal axis represents the count of surface objects in a single image, while the vertical axis indicates the number of images containing that specific count. Notably, the MEIWVD has the highest number of images (9,001) with 6 surface objects, followed by images with 5 and 4 objects. In contrast, the SeaShips dataset shows that 5,161 images contain a single surface object, accounting for 54.9\% of the total.

\ref{fig:dist_surface_objects} examines the width-to-height ratio of surface objects in both datasets. The horizontal axis represents the ratio of an object's width to its height, and the vertical axis depicts the number of surface objects. The data reveals a strong trend where surface objects generally have widths greater than heights, with most ratios concentrated between 3 and 5. Furthermore, the MEIWVD dataset, along with its annotation files, will be publicly released via a GitHub repository.

\section{The proposed object detection method}
\label{sec:sec4}
\begin{figure}[htbp]
    \centering
    % 插入图片
    \includegraphics[width=0.8\textwidth]{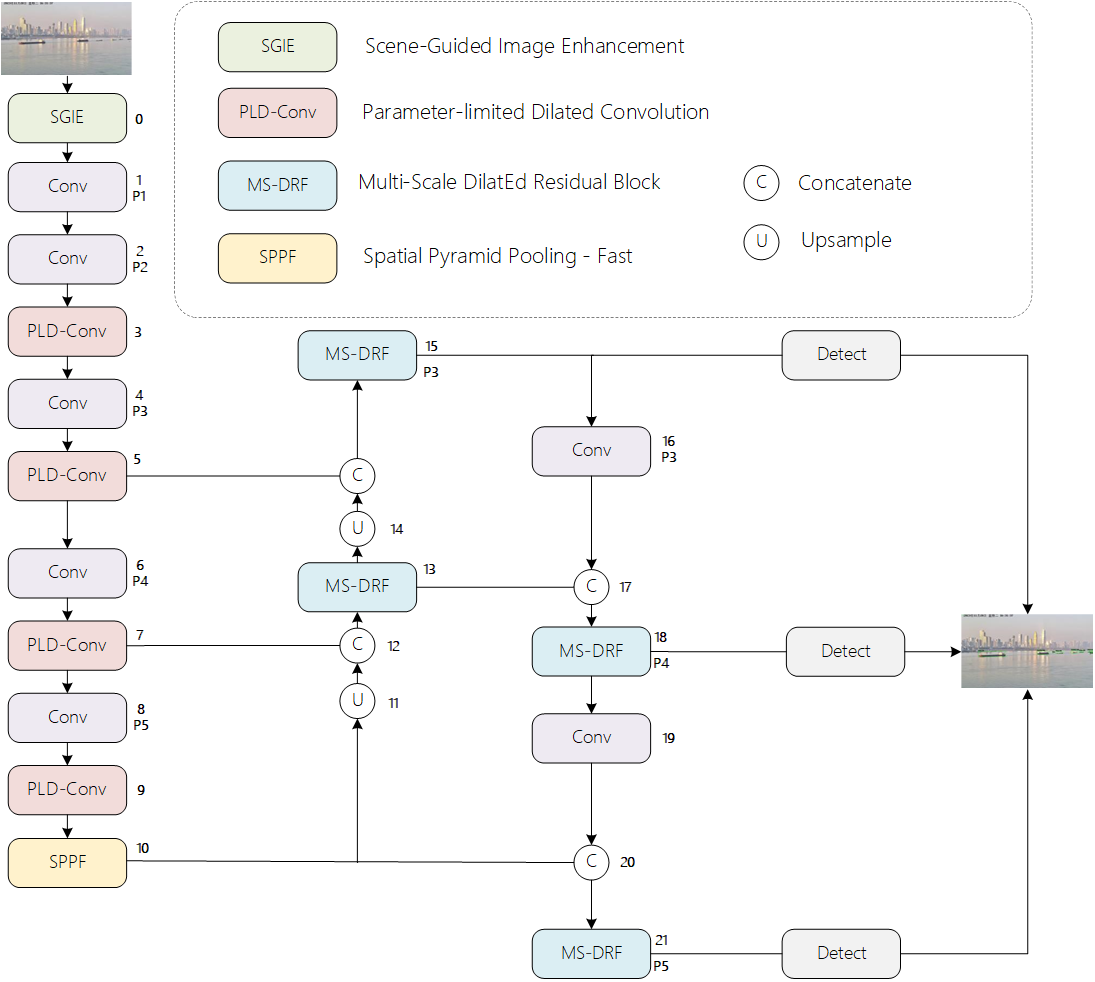} % 使用示例图片，替换为实际图片文件名
    % 图片标题
    \caption{The network architecture of multi-scene guided water surface object detection network (MSG-Net). The scene-guided image enhancement (SGIE) utilizes scene-guided embedding to generate contextual prompts, thereby effectively improving the quality of low-quality images. The parameter-limited dilated convolution (PLD-Conv) is suggested to extract feature information based on the geometric characteristics of water surface objects, while multi-scale dilated residual fusion (MS-DRF) is employed to hierarchically integrate feature information across different levels.}
    \label{fig:msg_net_architecture}
\end{figure}

To address the unique characteristics of MEIWVD, we propose a novel algorithm for water surface object detection in multi-environments which entitled multi-scene guided water surface object detection network (MSG-Net). The network architecture of MSG-Net is based on YOLOv8, which is illustrated in \ref{fig:msg_net_architecture}.

\subsection{Scene-guided image enhancement}
\label{sec:sec4.1}
The MEIWVD highlights the complexity of diverse environments and scenarios. Various factors can significantly affect water surface images, including water vapor, fog, and light, which can impair image quality to varying degrees. These factors not only affect the visual clarity of the images but also pose significant challenges to downstream object detection tasks. Therefore, improving image quality to enhance object detection performance has become an urgent and essential research direction.

While deep learning has achieved substantial advancements in mitigating image degradation---such as super-resolution, denoising, and deblurring---there remains a gap in adaptability to complex degradation scenarios, especially in tasks that necessitate flexible responses to multiple degradation types. Traditional methods typically involve identifying specific degradation types and applying corresponding enhancement strategies to improve visual quality \cite{guoD3NetIntegrated2023, liuAiOENetAllinOneLowVisibility2024}. This process often depends on widely used objective metrics, such as the structural similarity index (SSIM) and peak signal-to-noise ratio (PSNR).

In this context, existing water surface image enhancement methods often focus on specific scenarios (e.g., dehazing or deraining), which may lead to insufficient generalization capabilities due to scene variations. To address this issue, some researchers have proposed multi-task enhancement methods within a single network structure. For instance, AioENet improves visual perception of low-visibility images through a unified encoder-decoder network architecture \cite{liuAiOENetAllinOneLowVisibility2024}. CPA-Enhancer employs a chain-of-thought prompt adaptive enhancer to enhance object detection performance under unknown degradation conditions \cite{zhangCPAEnhancerChainofThoughtPrompted2025}. However, when processing images with similar content features but different degradation types, relying solely on content-based prompt generation strategies can be time-consuming, particularly in water surface monitoring scenarios.  

\begin{figure}[htbp]
    \centering
    % 插入图表
    \includegraphics[width=0.8\textwidth]{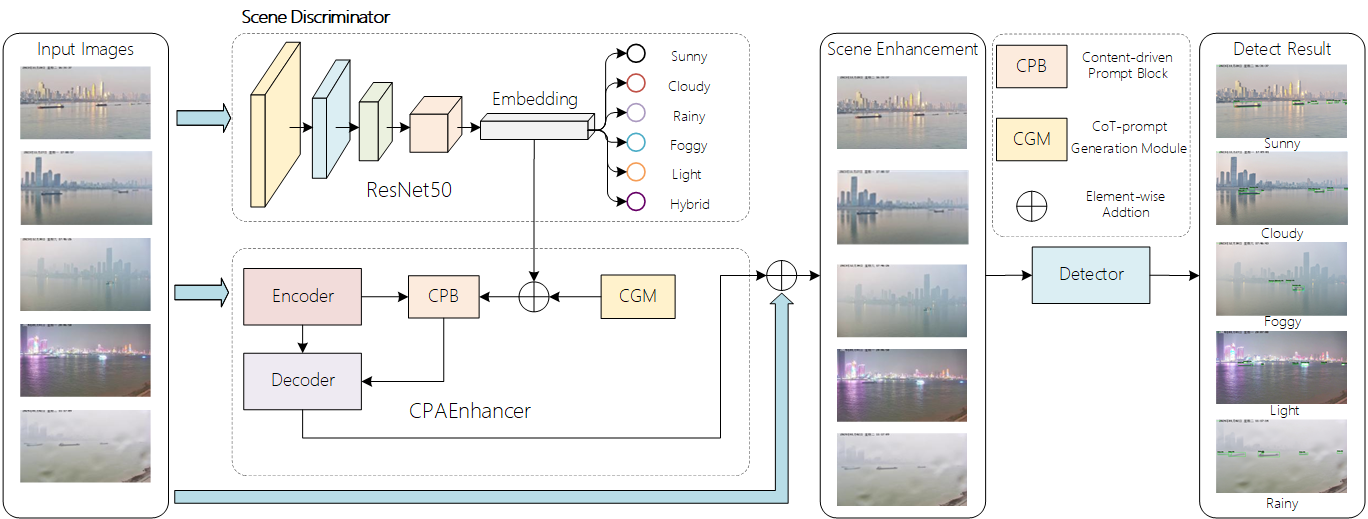} 
    % 图表标题
    \caption{Pipeline of the proposed scene-guided image enhancement (SGIE) module. SGIE extracts scene category features through the scene discriminator (SD) and integrates them with the prompts generated by the contextual guidance module (CGM) via vector addition.}
    \label{fig:sgie_pipeline}
\end{figure}

Inspired by CPA-Enhancer, we aim to bridge the gap between image enhancement and object detection tasks. CPA-Enhancer generates contextual prompts under completely unknown degradation conditions, which may affect the accuracy of prompt information during the initial stages of model training, especially in scenarios with multiple degradations, thereby impacting the performance of enhancement strategies. Considering the strong correlation between degradation types in water surface object detection datasets and quantifiable environmental parameters (e.g., visibility, rainfall intensity), we introduce a scene discriminator (SD) to extract scene category information, enhancing the ability to adapt to varying degradation types and improving overall performance in water surface object detection. The output feature vector embedding from SD is incorporated as guiding information into the prompt generator, leveraging known degradation types to guide the enhancement module for targeted improvements. Based on the above principles, we propose a scene-guided image enhancement (SGIE) module, which is illustrated in \ref{fig:sgie_pipeline}.

Specifically, SGIE first extracts scene category features through the scene discriminator (SD). By leveraging supervised learning, the scene category information effectively encapsulates the typical characteristics of the scene. These features are then vectorially added to the prompts generated by the contextual guidance module (CGM) and, together with the chain-of-thought prompt bank (CPB), are fed into the decoder to enhance the degraded images. CGM and CPB are the modules in CPA-Enhancer \cite{zhangCPAEnhancerChainofThoughtPrompted2025}. To reduce the computational complexity of the scene discriminator, its output is represented as a feature vector with a dimension of 512. The network framework of the scene discriminator adopts ResNet50 \cite{heDeepResidualLearning2016} as its backbone.  

The SGIE fully utilizes the category features of degradation labels, avoiding excessive reliance on image content features by the CPB module. This enables efficient and rapid guidance of the CPB to participate in the decoder's targeted enhancement of degraded images. 

\subsection{Parameter-limited dilated convolution}
\label{sec:sec4.2}
Existing deep learning-based object detection methods typically consist of three main components. Feature extraction module, feature fusion module, and the detection head \cite{redmonYOLOv3IncrementalImprovement2018}. The feature extraction module plays a crucial role as it is responsible for extracting rich feature information from the input image, which directly impacts the performance of subsequent object detection tasks. Traditional convolution operations extract features by sliding fixed-size convolution kernels over the input feature map, but their receptive fields are limited by the kernel size and stride. To overcome this limitation, dilated convolution significantly expands the receptive field by introducing gaps between the elements of the convolution kernel without increasing the number of parameters \cite{yuMultiscaleContextAggregation2016}. This method has been widely adopted in visual tasks such as semantic segmentation \cite{chenFrequencyAdaptiveDilatedConvolution2024,panDSSNetSimpleDilated2020,weiDWRSegRethinkingEfficient2022}. Additionally, deformable convolution allows the sampling points of the convolution kernel to dynamically adjust their positions based on the input data, enhancing the network's adaptability to geometric transformations[33,34]. Recently proposed snake-like dynamic convolution has demonstrated excellent performance in detecting slender and fragile tubular structures \cite{qiDynamicSnakeConvolution2023}.  

In the task of water surface object detection, common ship objects exhibit distinct geometric structures, typically characterized by regular rectangular shapes with relatively fixed width and height, where the width is significantly greater than the height, as shown in \ref{fig:aspect_ratio}. To enhance the flexibility of convolution operations and focus on the geometric features of water surface objects, inspired by snake-like dynamic convolution and deformable convolution, we propose a parameter-limited dilated convolution (PLD-Conv) based on geometric features. The goal is to efficiently capture the geometric characteristics of water surface objects by applying different constraints in various scale directions. 

\begin{figure}[htbp]
    \centering
    \includegraphics[width=\textwidth]{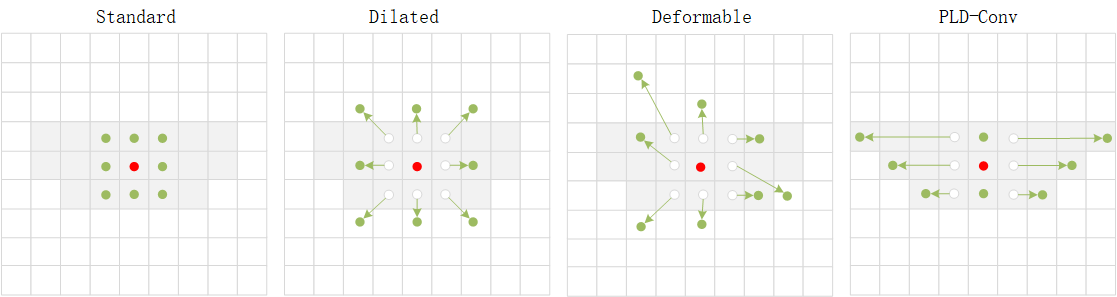} % 替换为实际图片文件名
    \vspace{0.5em} % 调整标题与标注之间的间距
    \begin{center}
        \small
        (a) Standard Conv \quad (b) Dilated Conv \quad (c) Deformable Conv \quad (d) PLD-Conv
    \end{center}
    \caption{Schematic diagrams of several typical convolution operations.}
    \label{fig:conv_operations}
\end{figure}

The core idea of PLD-Conv is to dynamically learn the offsets of input feature points in the width and height directions during the convolution process by introducing geometric constraints, thereby achieving direction-aware feature extraction. Specifically, PLD-Conv employs the Chebyshev distance as a metric to define the maximum coordinate difference between any two points \( p = (x_p, y_p) \) and \( q = (x_q, y_q) \) on the image, expressed as Eq. \eqref{eq1}.
\begin{equation}
    D_{\text{Chebyshev}}(p, q) = \max(|x_p - x_q|, |y_p - y_q|) \label{eq1}
\end{equation}
The Chebyshev distance is used to constrain the offset range of the convolution kernel sampling positions, ensuring that the sampling points remain within the local neighborhood and preventing excessive dispersion. Assuming the coordinates of an input feature point are \( (x, y) \), the sampling position after the convolution kernel offset \( P' \), where \( (\Delta x, \Delta y) \) is the offset dynamically learned by the network. To adapt to the rectangular characteristics of water surface targets, the offsets \( (\Delta x, \Delta y) \) must satisfy the constraints \( |\Delta x| \leq r_x \) and \( |\Delta y| \leq r_y \), where \( r_x \) and \( r_y \) are the median values of the aspect ratios of water surface targets (or thresholds set according to task requirements), constraining the offset ranges in the \( x \) and \( y \) directions, respectively.

\begin{equation}
P' = (x + \Delta x, y + \Delta y), \quad |\Delta x| \leq r_x, \quad |\Delta y| \leq r_y \label{eq2}
\end{equation}

\begin{equation}
D_C(P, P') \leq r, \quad P = (x, y), \quad P' = (x + \Delta x, y + \Delta y) \label{eq3}
\end{equation}
Additionally, the range of dynamic sampling must satisfy the Chebyshev distance constraint as Eq. \eqref{eq2}, where \( P \) is the center point of the convolution kernel, and \( r \) is the threshold for the Chebyshev distance. PLD-Conv adopts a bidirectional strategy, using different convolution operations in the \( x \) and \( y \) directions: in the \( x \) direction, dilated convolution is employed to rapidly expand the receptive field horizontally, expressed as Eq. \eqref{eq4}, where \( I \) is the input feature map, \( k_x \) is the convolution kernel, and \( d \) is the dilation rate. In the \( y \) direction, standard convolution is used to extract relevant features, expressed as Eq. \eqref{eq5}, where \( k_y \) is the convolution kernel.

\begin{equation}
    F_x = \text{DilatedConv}(I, k_x, d) \label{eq4}
\end{equation}

\begin{equation}
    F_y = \text{Conv}(I, k_y) \label{eq5}
\end{equation}

\begin{equation}
    F_{\text{PLD}} = \text{PLD-Conv}(I, k_x, k_y, \Delta x, \Delta y, r) \label{eq6}
\end{equation}
Through this bidirectional strategy, PLD-Conv can effectively adapt to the rectangular characteristics of water surface targets, enhancing the model’s geometric perception and feature extraction accuracy. Combining the above formulas, the overall operation of PLD-Conv can be expressed as Eq. \eqref{eq6}. By introducing geometric constraints and a bidirectional convolution strategy, PLD-Conv can flexibly capture local neighborhood features while limiting the perceptual range, enabling the convolution kernel to focus more on the structural characteristics of water surface targets, thereby improving the model’s performance. \ref{fig:conv_operations} compares the receptive field ranges of several typical convolution operations.

\subsection{Multi-scale dilated residual fusion}
\label{sec:sec4.3}
Through statistical analysis of the MEIWVD (refer to Section 3.2), it is observed that water surface objects exhibit dual complex characteristics: multi-scale representation specificity and spatial distribution density. These characteristics pose significant challenges for object detection algorithms. Key objects such as ships and buoys often occupy less than 1\% of the total pixels (typical sizes around 20\texttimes10 pixels), making them prone to losing shallow texture and edge features during the forward propagation of conventional convolutional neural networks. Additionally, due to differences in shooting angles and object distances, objects of the same category exhibit significant scale variations in images. For instance, the scale difference between nearby and distant ships can exceed 5 times, making it difficult for detectors with a single receptive field to effectively capture features of objects at varying distances. Furthermore, in inland river scenes, objects are often densely distributed, with a single frame frequently containing multiple closely spaced objects. Overlapping between objects is common, leading to semantic interference in the feature map and resulting in false detections and missed detections.  

\begin{figure}[htbp]
    \centering
    \includegraphics[width=0.5\textwidth,height=0.5\textheight]{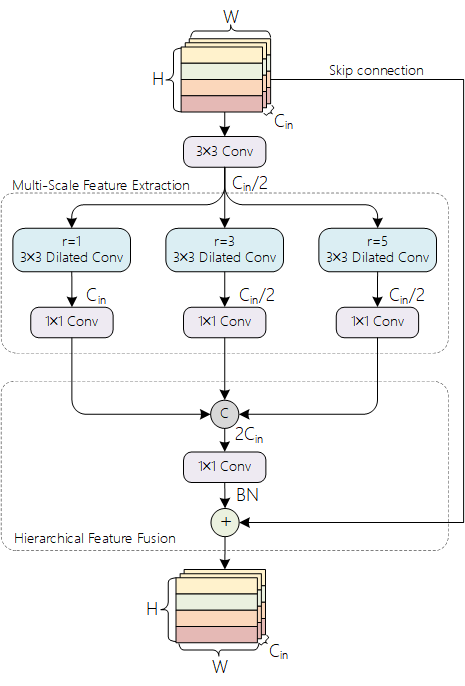} % 替换为实际图片文件名
    \caption{Multi-scale dilated residual fusion module.}
    \label{fig:multi_scale_dilated_residual_fusion}
\end{figure}

To address these issues, we propose an innovative multi-scale feature fusion module named multi-scale dilated residual fusion (MS-DRF), which is illustrated in \ref{fig:multi_scale_dilated_residual_fusion}. Inspired by depthwise separable convolution, MS-DRF adopts a strategy of feature extraction with varying dilation rates and hierarchical feature fusion to achieve multi-scale feature integration. Firstly, the input feature map undergoes a 3\texttimes3 convolution to compress the feature length, reducing the computational cost of multi-layer dilated convolution operations. Subsequently, convolution kernels with different dilation rates are stacked in parallel, progressively expanding the receptive field to form multiple incremental scale-aware mechanisms. To avoid noise interference caused by simple concatenation of multi-scale features, the results of these incremental scale-aware operations are individually fused through 1\texttimes1 convolutions before concatenation. To effectively utilize information from multiple scales, global features are fused through a 1\texttimes1 convolution, and skip connections are employed to enhance the original information, enabling deep propagation.

To elaborate on the specific process of MS-DRF, assume three dilation rates (r = 1, 3, 5) are used for progressive dilated convolution. The MS-DRF process is described by Eq. \eqref{eq7} to Eq. \eqref{eq10}. Eq. \eqref{eq7} compresses the feature length of the feature map using a 3\texttimes3 convolution. Eq. \eqref{eq8} extracts multi-scale features through dilated convolutions with different dilation rates. To better preserve the feature information of small objects, the output feature length for the dilation rate r = 1 is twice that of the other dilated convolutions. In Eq. \eqref{eq9},  we fuse, concatenate, and integrate multiple progressive scales, effectively combining multi-scale features. To mitigate information attenuation caused by the hierarchical structure, Eq. \eqref{eq10} employs residual connections to pass local and global features to deeper layers through residual paths.

\begin{equation}
    F_0 = \text{Conv}_{3 \times 3}(F_{\text{input}}, W_0, b_0) 
    \label{eq7}
\end{equation}
    
\begin{equation}
    F_i = \text{DilatedConv}_{3 \times 3, r = i}(F_0, W_i, b_i) 
    \label{eq8}
\end{equation}

\begin{equation}
    F_{\text{fused}} = \text{BN}(\text{Concat}(\text{Conv}_{1 \times 1}(F_i, W_i, b_i))) 
    \label{eq9}
\end{equation}
    
\begin{equation}
    F_{\text{out}} = F_{\text{fused}} + \text{Conv}_{1 \times 1}(F_{\text{in}}, W, b) 
    \label{eq10}
\end{equation}
       
\section{Experimental results}
\label{sec:sec5}
In this section, we conduct benchmark validation and in-depth analysis on MEIWVD. By employing state-of-the-art deep learning-based object detection models, we systematically evaluating the models' performance across various scenarios and complex weather conditions. Additionally, to validate the effectiveness of the proposed method MSG-Net, we conduct a comparative analysis of performance enhancements achieved through improvements in multi-scenario adaptability, geometric feature extraction, and multi-scale characterization of water surface objects, aiming to identify potential optimization strategies for inland waterway vessel detection in real-world environments.  

\subsection{Datasets}
\label{sec:sec5.1}
To validate the effectiveness of the proposed method, experiments are conducted on two datasets: the publicly available SeaShips \cite{shaoSeaShipsLargeScalePrecisely2018} dataset and our newly constructed MEIWVD. The SeaShips dataset contains object detection samples of various maritime vessels, while the MEIWVD focuses on inland waterway scenarios, encompassing a rich collection of real-world data under diverse environmental conditions and multi-scale objects. This dual-validation strategy enables a comprehensive evaluation of the proposed method's applicability and robustness from the perspectives of different scenarios and data distributions.

\subsection{Experimental details}
\label{sec:sec5.2}
The MSG-Net architecture we constructed is based on the foundational framework of YOLOv8. All experiments are conducted in the PyTorch deep learning environment, utilizing an RTX 2080Ti GPU as the computational platform. The hyper parameters are configured as follows: The optimizer employs stochastic gradient descent (SGD) with an initial learning rate set to 1e-4, which is dynamically adjusted during training based on the loss. The batch size is set to 16 and the training process spans 100 epochs. To ensure the model's stability and convergence, we meticulously recorded and compared the performance at various stages of training, aiming to validate the model's generalization capabilities in complex scenarios. 

\subsection{Experimental results and analysis}
\label{sec:sec5.3}

\subsubsection{Ablation experiments}
\label{sec:sec5.3.1}
To analyze the contributions of the three modules in MSG-Net object detection performance, we conducted corresponding ablation experiments. Table \ref{tab:module_validation} presents the performance of the three proposed modules on SeaShips and MEIWVD. We adopt YOLOv8 as the baseline model.

The SGIE module focuses on scene-guided prompt generation to improve multi-scenario object detection performance. From Table \ref{tab:module_validation}, we can notice that, on the SeaShips, which lacks specific degradation category information and consists solely of high-resolution images, experimental results indicate that the SGIE module leads to a slight performance degradation, with a 0.4\% decrease in accuracy. However, on the MEIWVD, which includes multi-scenario and complex degradation environments, the SGIE module successfully improves detection accuracy by 0.9\% increase through its scene-guided triggering mechanism. This result highlights the importance of matching dataset characteristics with enhancement strategies.  The PLD-Conv demonstrates significant improvements in water surface object detection by optimizing feature extraction. Experimental results in Table \ref{tab:module_validation} reveal a 1.4\% enhancement in mAP\text@[0.5:0.95] on the SeaShips and a 0.9\% improvement on the MEIWVD, attributed to its capability to enhance the structured features of water surface objects. The MS-DRF module addresses multi-scale object detection through dynamic feature fusion, effectively improving the model's adaptability to objects of varying sizes. Performance evaluations in Table \ref{tab:module_validation} show a 1.5\% increase in mAP\text@[0.5:0.95] on the SeaShips and a 0.7\% improvement on the MEIWVD, confirming its universal applicability and effectiveness across diverse data distributions. Comparative experiments on the two datasets demonstrate that the three proposed modules collectively enhance the detection performance of water surface objects. 

\begin{table}[htbp]
	\centering
	\caption{Validation of the effectiveness of the three modules.}
	\begin{tabular}{p{3cm}p{3cm}p{3cm}}
		\toprule
		\multirow{2}[4]{*}{} & \multicolumn{2}{c}{mAP\text@[0.5:0.95]} \\
		\cmidrule{2-3}          & \multicolumn{1}{c}{SeaShips} & \multicolumn{1}{c}{MEIWVD} \\
		\midrule
		Baseline & 78.5\% & 80.2\% \\
		\midrule
		+ SGIE & 78.1\% (-0.4\%) & 81.1\% (+0.9\%) \\
		+ PLD-Conv & 79.9\% (+1.4\%) & 81.1\% (+0.9\%) \\
		+ MS-DRF & 80.0\% (+1.5\%) & 80.9\% (+0.7\%) \\
		\bottomrule
	\end{tabular}%
	\label{tab:module_validation}%
\end{table}%

\subsubsection{Ablation study on method combinations}
\label{sec:sec5.3.2}
To further explore the synergistic effects of the three proposed modules, we designed multiple ablation experiments to validate their contributions to the final performance through different combinations. Since the SeaShips dataset does not represent a multi-scenario, image enhancement on the dataset led to a decline in object detection performance. Therefore, the ablation experiments for method combinations primarily focused on the MEIWVD, which includes multi-scenarios. The specific experimental results are presented in Table \ref{tab:ablation_study}.

The integration of SGIE and MS-DRF enhances performance by 0.8\% over the baseline model, indicating the effectiveness of scene-guided image enhancement and multi-scale processing in addressing its limitations in complex scenarios. The integration of the SGIE module and PLD-Conv yields a 1.1\% performance improvement, demonstrating the complementary relationship between water surface object feature extraction and multi-scenario enhancement. Furthermore, the combination of PLD-Conv and the MS-DRF module achieves better results, with improvements of 1.2\%, highlighting the synergistic effect of multi-scale feature fusion in enhancing detection accuracy. Finally, the combined application of all three methods results in a 1.4\% improvement. The experimental results demonstrate that the three proposed modules significantly improve object detection performance. Specifically, the SGIE module achieves targeted improvements by addressing diverse environmental conditions in multi-scenario enhancement. The PLD-Conv module effectively addresses the unique shape characteristics of water surface objects, validating its robust feature extraction capability, while the MS-DRF module demonstrates strong adaptability in handling multi-scale objects. 

% \begin{table}[htbp]
% 	\centering
% 	\caption{Ablation study results of module combinations on the MEIWVD.}
% 	\begin{tabular}{cccc}
% 		\toprule
% 		SGIE & PLD-Conv & MS-DRF & mAP\text@[0.5:0.95] \\
% 		\midrule
% 		& & & 80.2\% \\
% 		\checkmark & & \checkmark & 81.0\% (+0.8\%) \\
% 		\checkmark & \checkmark & & 81.3\% (+1.1\%) \\
% 		& \checkmark & \checkmark & 81.4\% (+1.2\%) \\
% 		\checkmark & \checkmark & \checkmark & \textbf{81.6\%} \textbf{(+1.4\%)} \\
% 		\bottomrule
% 	\end{tabular}%
% 	\label{tab:ablation_study}%
% \end{table}%

\begin{table}[htbp]
	\centering
	\caption{Ablation study results of module combinations on the MEIWVD.}
	\begin{tabular}{ccccc}
		\toprule
		\textbf{Method} & \textbf{SGIE} & \textbf{PLD-Conv} & \textbf{MS-DRF} & \textbf{mAP\text@[0.5:0.95]} \\
		\midrule
		Baseline & & & & 80.2\% \\
		1 & \checkmark & & \checkmark & 81.0\% (+0.8\%) \\
		2 & \checkmark & \checkmark & & 81.3\% (+1.1\%) \\
		3 & & \checkmark & \checkmark & 81.4\% (+1.2\%) \\
		4 & \checkmark & \checkmark & \checkmark & \textbf{81.6\%} \textbf{(+1.4\%)} \\
		\bottomrule
	\end{tabular}%
	\label{tab:ablation_study}%
\end{table}%

\subsubsection{Comparative experiments with state-of-the-art methods}
\label{sec:sec5.3.3}
To evaluate the detection performance of MSG-Net on both datasets, we conducted comparative experiments with several widely-used and advanced object detection algorithms, including DETR \cite{carionEndtoEndObjectDetection2020a}, Deformable DETR \cite{zhuDeformableDetrDeformable2021}, YOLOv8, YOLOv11 \cite{khanamYOLOv11OverviewKey2024}. The experimental results are presented in Table \ref{tab:performance_comparison}.  As demonstrated in Table \ref{tab:performance_comparison}, MSG-Net surpasses existing object detection methods on both the SeaShips and MEIWVD, achieving mAP scores of 80.7\% and 81.6\% respectively under the [0.5:0.95] IoU threshold. On SeaShips, MSG-Net outperforms DETR, Deformable DETR, YOLOv8, and YOLOv11 by 42.1\%, 24.9\%, 2.8\%, and 1.1\% respectively. Similarly, on MEIWVD, it achieves a mAP of 81.6\%, exceeding YOLOv8 (80.2\%) and YOLOv11 (80.3\%). The result demonstrates that MSG-Net exhibits stronger generalization capabilities and higher accuracy in detecting water surface objects across multiple scenarios. MSG-Net's superior performance stems from its optimized architecture, which integrates multi-scenario adaptability, enhanced water surface object feature extraction, and effective multi-scale feature fusion, demonstrating significant advantages over existing algorithms in high-precision object detection tasks.

\begin{table}[htbp]
	\centering
	\caption{Performance comparison of object detection methods on maritime datasets.}
	\begin{tabular}{lcc}
		\toprule
		\multirow{2}[4]{*}{Method} & \multicolumn{2}{c}{mAP\text@[0.5:0.95]} \\
		\cmidrule{2-3}          & \multicolumn{1}{c}{SeaShips} & \multicolumn{1}{c}{MEIWVD} \\
		\midrule
		DETR & 56.8\% & 50.2\% \\
		Deformable DETR & 64.6\% & 59.4\% \\
		YOLOv8 & 78.5\% & 80.2\% \\
		YOLOv11 & 79.8\% & 80.3\% \\
		MSG-Net (Ours) & \textbf{80.7\%} (no SGIE) & \textbf{81.6\%} \\
		\bottomrule
	\end{tabular}%
	\label{tab:performance_comparison}%
\end{table}%

\subsubsection{Qualitative detection results}
\label{sec:sec5.3.4}
This section provides a visual representation of the detection results to qualitatively assess the performance of the proposed MSG-Net. The qualitative analysis focuses on the model's ability to accurately detect and localize objects under various conditions, including complex weather, multi-scale scenarios, and diverse environmental settings. To clearly display the detection results, abbreviations are used to represent categories in the images, such as CG for cargo ship, CS for container ship, PS for passenger ship, and BY for buoy, ensuring clarity and conciseness in the visualization.

\begin{figure}[htbp]
    \centering
    \includegraphics[width=\textwidth]{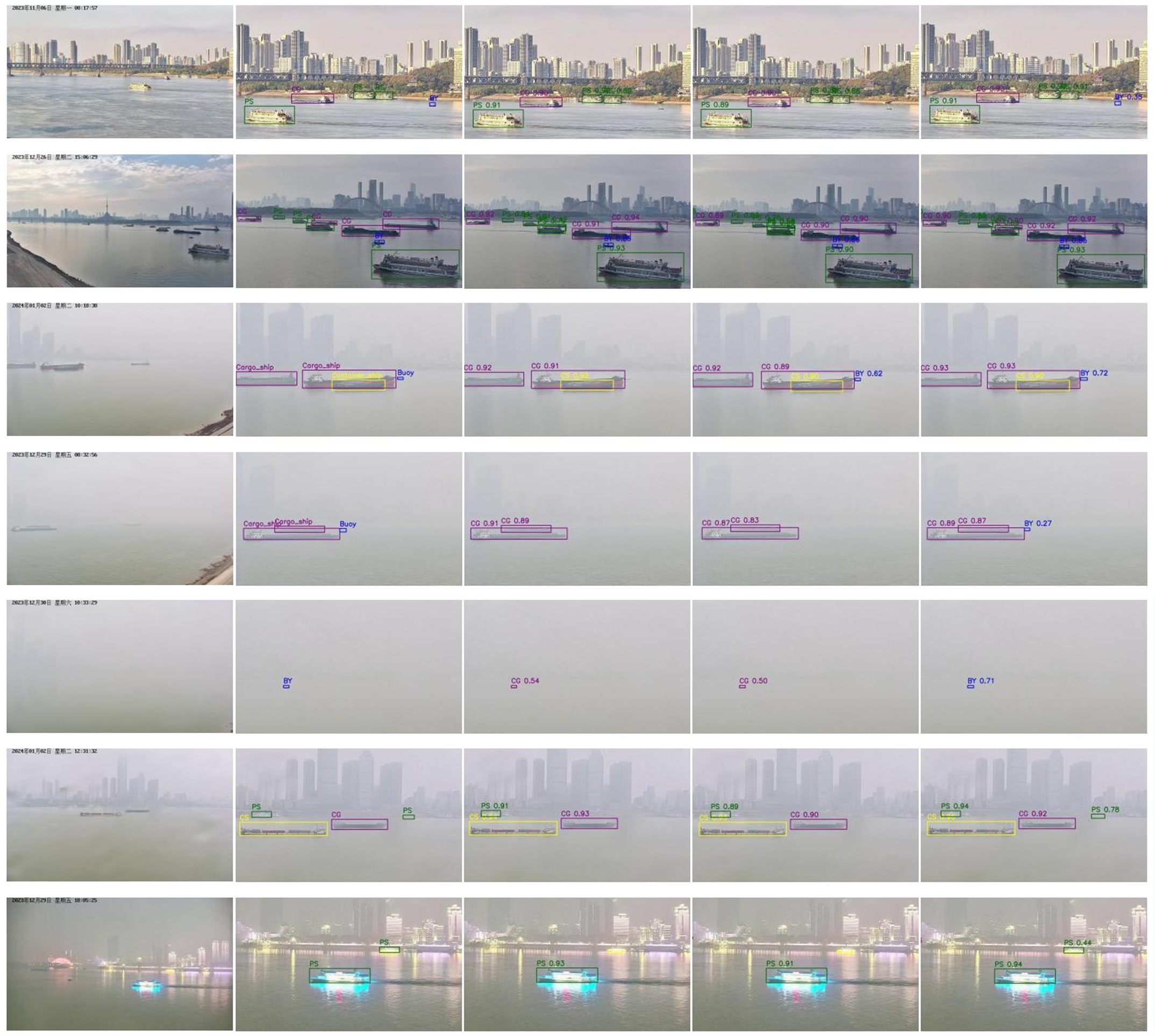} % 替换为你的图片文件名
    \begin{center}
        \footnotesize
        (a) Original image  \quad  (b) Ground truth  \quad  (c) YOLOv8  \quad (d) YOLOv11  \quad  (e) MSG-Net
    \end{center}
    \vspace{-0.5em} % 调整标题与标注之间的间距
    \caption{The comparisons of surface object detection result across multi-environmental images. The figure shows the detection results obtained by YOLOv8n, YOLOv11n, and MSG-Net, highlighting the performance of each algorithm in detecting surface objects under varying environmental conditions. MSG-Net demonstrates superior accuracy and robustness in handling complex scenarios.}
    \label{fig:detection_results}
\end{figure}

In this section, we randomly select some images and compared the results of the compared methods. Due to the high resolution of the images and the relatively small proportion of ship objects on the water surface, we cropped the key regions to highlight the detection results for clearer presentation. The experimental scenarios included various complex environments such as sunny, cloudy, moderate fog, dense fog, rainy, and mixed artificial lighting with thin fog conditions.

As shown in \ref{fig:detection_results}, in the first row (sunny scenario), despite favorable weather conditions, small objects (e.g., buoys) are still missed by YOLOv8 and YOLOv11. This is due to the strong sunlight reflection on the water surface, which causes the features of small objects to be confused with the lighting information. In contrast, MSG-Net significantly improves the detection of small objects through environmental enhancement and feature fusion. In the second row (cloudy scenario), YOLOv8 and YOLOv11 misidentify a cargo ship as a passenger ship in a dense ships' scene. This is attributed to the confusion of features caused by the depth-of-field effect. MSG-Net, with its specialized feature fusion, accurately identify the distant ship of the same type and provided higher confidence scores. In the third row (moderate fog scenario), due to reduced visibility of small objects, YOLOv8 failed to detect the buoy, while YOLOv11 and MSG-Net successfully detect them. MSG-Net demonstrated higher confidence in detecting small objects. In the fourth and fifth rows (dense fog scenarios), YOLOv8 and YOLOv11 either missed the buoy or misidentified it as a Cargo Ship. MSG-Net, enhanced by SGIE, accurately identified the buoy category, showcasing its robustness in challenging scenarios. In the sixth row (rainy scenario), YOLOv8 and YOLOv11 missed the Passenger Ship. MSG-Net effectively addressed this issue through environmental feature enhancement. In the last row (mixed artificial lighting and thin fog scenario), the detectors often misidentified a lit yacht as shore lighting, leading to missed detections. MSG-Net accurately identified the Passenger Ship docked near the shore, demonstrating its superior performance under complex lighting conditions.

\section{Conclusions}
In this paper, we introduced the multi-environment inland water vessel detection (MEIWVD) dataset, a foundational resource for researching vessel object detection in complex inland river environments. We detailed the dataset's construction process, including data collection, annotation, and classification standards, and conducted an in-depth analysis of its characteristics to highlight its advantages in multi-environment and multi-scale scenarios.

To address the dataset's unique features, we proposed a series of improvements focusing on three aspects: enhancement and adaptation to multi-environment characteristics, feature extraction for water surface objects, and fusion and processing of multi-scale features. These methods not only improved the model's detection performance under diverse environmental conditions but also provided new insights and methodologies for future research in similar scenarios.  However, despite the achievements of this study, several potential research directions warrant further exploration. First, the impact of varying lighting conditions, such as day-night transitions and natural versus artificial lighting, on object detection remains to be thoroughly investigated. Second, the current dataset is limited to the Yangtze River basin, and the diversity of vessel types is constrained by the characteristics of this region. Expanding the dataset to encompass more diverse inland river environments and vessel types will be a critical focus of future research.  Furthermore, while the MEIWVD provides a solid foundation for inland water object detection, we hope it will inspire more researchers to delve into this field and address the diverse challenges of real-world scenarios. We plan to continue expanding and refining the dataset, exploring its broader application potential, and fostering the advancement of inland water object detection technology through collaboration and open data sharing. 

\section*{Authorship Contribution Statement}
\textbf{Shanshan Wang:} Conceptualization, Methodology, Writing – original draft. \textbf{Haixiang Xu:} Conceptualization, Methodology, Writing – review \& editing. \textbf{Hui Feng:} Conceptualization, Methodology, Writing – review \& editing. \textbf{Xiaoqian Wang:} Investigation, Resources. \textbf{Pei Song:}  Data Curation. \textbf{Sijie Liu:} Data Curation. \textbf{Jianhua He:} Writing – review \& editing.

\section*{Declaration of competing interest}
The authors declare that they have no known competing financial interests or personal relationships that could have appeared to influence the work reported in this paper.

\section*{Acknowledgement}
The authors appreciate the constructive suggestions from reviewers and the Associate Editor. This work is supported by the National Natural Science Foundation of China under Grant No.52371374, 51979210. This work was partly funded by EPSRC with RC Grant reference EP/Y027787/1, UKRI under grant number EP/Y028317/1, the Horizon Europe MSCA programme under grant agreement No 101086228.



\bibliographystyle{elsarticle-harv} 
\bibliography{mybib}

\begin{thebibliography}{52}
\expandafter\ifx\csname natexlab\endcsname\relax\def\natexlab#1{#1}\fi
\providecommand{\url}[1]{\texttt{#1}}
\providecommand{\href}[2]{#2}
\providecommand{\path}[1]{#1}
\providecommand{\DOIprefix}{doi:}
\providecommand{\ArXivprefix}{arXiv:}
\providecommand{\URLprefix}{URL: }
\providecommand{\Pubmedprefix}{pmid:}
\providecommand{\doi}[1]{\href{http://dx.doi.org/#1}{\path{#1}}}
\providecommand{\Pubmed}[1]{\href{pmid:#1}{\path{#1}}}
\providecommand{\bibinfo}[2]{#2}
\ifx\xfnm\relax \def\xfnm[#1]{\unskip,\space#1}\fi
%Type = Article
\bibitem[{Arkin et~al.(2023)Arkin, Yadikar, Xu, Aysa and Ubul}]{arkinSurveyObjectDetection2023}
\bibinfo{author}{Arkin, E.}, \bibinfo{author}{Yadikar, N.}, \bibinfo{author}{Xu, X.}, \bibinfo{author}{Aysa, A.}, \bibinfo{author}{Ubul, K.}, \bibinfo{year}{2023}.
\newblock \bibinfo{title}{A survey: object detection methods from {CNN} to transformer}.
\newblock \bibinfo{journal}{Multimedia Tools and Applications} \bibinfo{volume}{82}, \bibinfo{pages}{21353--21383}.
%Type = Article
\bibitem[{Bochkovskiy et~al.(2020)Bochkovskiy, Wang and Liao}]{bochkovskiyYOLOv4OptimalSpeed2020}
\bibinfo{author}{Bochkovskiy, A.}, \bibinfo{author}{Wang, C.Y.}, \bibinfo{author}{Liao, H.Y.M.}, \bibinfo{year}{2020}.
\newblock \bibinfo{title}{Yolov4: Optimal {Speed} and {Accuracy} of {Object} {Detection}}.
\newblock \bibinfo{journal}{arXiv preprint} \bibinfo{note}{ArXiv:2004.10934}.
%Type = Article
\bibitem[{Cai et~al.(2024)Cai, Meng and Wu}]{caiFEYOLOYOLOShip2024}
\bibinfo{author}{Cai, S.}, \bibinfo{author}{Meng, H.}, \bibinfo{author}{Wu, J.}, \bibinfo{year}{2024}.
\newblock \bibinfo{title}{Fe-{YOLO}: Yolo {Ship} {Detection} {Algorithm} based on {Feature} {Fusion} and {Feature} {Enhancement}}.
\newblock \bibinfo{journal}{Journal of Real-Time Image Processing} \bibinfo{volume}{21}, \bibinfo{pages}{61}.
%Type = Article
\bibitem[{Cao et~al.(2022)Cao, Yuan, Feng and Niu}]{caoCFDETRCoarsetoFineTransformers2022}
\bibinfo{author}{Cao, X.}, \bibinfo{author}{Yuan, P.}, \bibinfo{author}{Feng, B.}, \bibinfo{author}{Niu, K.}, \bibinfo{year}{2022}.
\newblock \bibinfo{title}{Cf-{DETR}: Coarse-to-{Fine} {Transformers} for {End}-to-{End} {Object} {Detection}}.
\newblock \bibinfo{journal}{Proceedings of the AAAI Conference on Artificial Intelligence} \bibinfo{volume}{36}, \bibinfo{pages}{185--193}.
%Type = Article
\bibitem[{Carion et~al.(2020)Carion, Massa, Synnaeve, Usunier, Kirillov and Zagoruyko}]{carionEndtoEndObjectDetection2020a}
\bibinfo{author}{Carion, N.}, \bibinfo{author}{Massa, F.}, \bibinfo{author}{Synnaeve, G.}, \bibinfo{author}{Usunier, N.}, \bibinfo{author}{Kirillov, A.}, \bibinfo{author}{Zagoruyko, S.}, \bibinfo{year}{2020}.
\newblock \bibinfo{title}{End-to-{End} {Object} {Detection} with {Transformers}}.
\newblock \bibinfo{journal}{Proceedings of the {European} {Conference}} , \bibinfo{pages}{213--229}.
%Type = Article
\bibitem[{Chen et~al.(2024)Chen, Gu, Zheng and Fu}]{chenFrequencyAdaptiveDilatedConvolution2024}
\bibinfo{author}{Chen, L.}, \bibinfo{author}{Gu, L.}, \bibinfo{author}{Zheng, D.}, \bibinfo{author}{Fu, Y.}, \bibinfo{year}{2024}.
\newblock \bibinfo{title}{Frequency-{Adaptive} {Dilated} {Convolution} for {Semantic} {Segmentation}}.
\newblock \bibinfo{journal}{Computer {Vision} and {Pattern} {Recognition} ({CVPR})} , \bibinfo{pages}{3414--3425}.
%Type = Article
\bibitem[{Dai et~al.(2017)Dai, Qi, Xiong, Li, Zhang, Hu and Wei}]{daiDeformableConvolutionalNetworks2017a}
\bibinfo{author}{Dai, J.}, \bibinfo{author}{Qi, H.}, \bibinfo{author}{Xiong, Y.}, \bibinfo{author}{Li, Y.}, \bibinfo{author}{Zhang, G.}, \bibinfo{author}{Hu, H.}, \bibinfo{author}{Wei, Y.}, \bibinfo{year}{2017}.
\newblock \bibinfo{title}{Deformable {Convolutional} {Networks}}.
\newblock \bibinfo{journal}{Proceedings of the {IEEE} {International} {Conference} on {Computer} {Vision}} , \bibinfo{pages}{764--773}.
%Type = Article
\bibitem[{Dalal and Triggs(2005)}]{dalalHistogramsOrientedGradients2005}
\bibinfo{author}{Dalal, N.}, \bibinfo{author}{Triggs, B.}, \bibinfo{year}{2005}.
\newblock \bibinfo{title}{Histograms of {Oriented} {Gradients} for {Human} {Detection}}.
\newblock \bibinfo{journal}{Proceedings of {IEEE} {Computer} {Society} {Conference} on {Computer} {Vision} and {Pattern} {Recognition}} , \bibinfo{pages}{886--893}.
%Type = Article
\bibitem[{Er et~al.(2023)Er, Zhang, Chen and Gao}]{erShipDetectionDeep2023}
\bibinfo{author}{Er, M.J.}, \bibinfo{author}{Zhang, Y.}, \bibinfo{author}{Chen, J.}, \bibinfo{author}{Gao, W.}, \bibinfo{year}{2023}.
\newblock \bibinfo{title}{Ship detection with deep learning: a survey}.
\newblock \bibinfo{journal}{Artificial Intelligence Review} \bibinfo{volume}{56}, \bibinfo{pages}{11825--11865}.
%Type = Article
\bibitem[{Feng et~al.(2021)Feng, Guo, Xu and Ge}]{fengSharpGANDynamicScene2021}
\bibinfo{author}{Feng, H.}, \bibinfo{author}{Guo, J.}, \bibinfo{author}{Xu, H.}, \bibinfo{author}{Ge, S.S.}, \bibinfo{year}{2021}.
\newblock \bibinfo{title}{Sharpgan: Dynamic {Scene} {Deblurring} {Method} for {Smart} {Ship} {Based} on {Receptive} {Field} {Block} and {Generative} {Adversarial} {Networks}}.
\newblock \bibinfo{journal}{Sensors} \bibinfo{volume}{21}, \bibinfo{pages}{3641}.
%Type = Article
\bibitem[{Girshick(2015)}]{girshickFastRCNN2015a}
\bibinfo{author}{Girshick, R.}, \bibinfo{year}{2015}.
\newblock \bibinfo{title}{Fast {R}-{CNN}}.
\newblock \bibinfo{journal}{Preceedings of {IEEE} {International} {Conference} on {Computer} {Vision}} , \bibinfo{pages}{1440--1448}.
%Type = Article
\bibitem[{Girshick et~al.(2016)Girshick, Donahue, Darrell and Malik}]{girshickRegionbasedConvolutionalNetworks2016}
\bibinfo{author}{Girshick, R.}, \bibinfo{author}{Donahue, J.}, \bibinfo{author}{Darrell, T.}, \bibinfo{author}{Malik, J.}, \bibinfo{year}{2016}.
\newblock \bibinfo{title}{Region-based {Convolutional} {Networks} for {Accurate} {Object} {Detection} and {Segmentation}}.
\newblock \bibinfo{journal}{IEEE Transactions on Pattern Analysis and Machine Intelligence} \bibinfo{volume}{38}, \bibinfo{pages}{142--158}.
%Type = Article
\bibitem[{Guo et~al.(2020)Guo, Yang, Wang, Song and Gao}]{guoRotationalLibraRCNN2020}
\bibinfo{author}{Guo, H.}, \bibinfo{author}{Yang, X.}, \bibinfo{author}{Wang, N.}, \bibinfo{author}{Song, B.}, \bibinfo{author}{Gao, X.}, \bibinfo{year}{2020}.
\newblock \bibinfo{title}{A {Rotational} {Libra} {R}-{CNN} {Method} for {Ship} {Detection}}.
\newblock \bibinfo{journal}{IEEE Transactions on Geoscience and Remote Sensing} \bibinfo{volume}{58}, \bibinfo{pages}{5772--5781}.
%Type = Article
\bibitem[{Guo et~al.(2023)Guo, Feng, Xu, Yu and Ge}]{guoD3NetIntegrated2023}
\bibinfo{author}{Guo, J.}, \bibinfo{author}{Feng, H.}, \bibinfo{author}{Xu, H.}, \bibinfo{author}{Yu, W.}, \bibinfo{author}{Ge, S.S.}, \bibinfo{year}{2023}.
\newblock \bibinfo{title}{D3 -{Net}: Integrated multi-task convolutional neural network for water surface deblurring, dehazing and object detection}.
\newblock \bibinfo{journal}{Engineering Applications of Artificial Intelligence} \bibinfo{volume}{117}, \bibinfo{pages}{105558}.
%Type = Article
\bibitem[{He et~al.(2015)He, Zhang, Ren and Sun}]{heSpatialPyramidPooling2015}
\bibinfo{author}{He, K.}, \bibinfo{author}{Zhang, X.}, \bibinfo{author}{Ren, S.}, \bibinfo{author}{Sun, J.}, \bibinfo{year}{2015}.
\newblock \bibinfo{title}{Spatial {Pyramid} {Pooling} in {Deep} {Convolutional} {Networks} for {Visual} {Recognition}}.
\newblock \bibinfo{journal}{IEEE Transactions on Pattern Analysis and Machine Intelligence} \bibinfo{volume}{37}, \bibinfo{pages}{1904--1916}.
%Type = Article
\bibitem[{He et~al.(2016)He, Zhang, Ren and Sun}]{heDeepResidualLearning2016}
\bibinfo{author}{He, K.}, \bibinfo{author}{Zhang, X.}, \bibinfo{author}{Ren, S.}, \bibinfo{author}{Sun, J.}, \bibinfo{year}{2016}.
\newblock \bibinfo{title}{Deep {Residual} {Learning} for {Image} {Recognition}}.
\newblock \bibinfo{journal}{2016 {IEEE} {Conference} on {Computer} {Vision} and {Pattern} {Recognition} ({CVPR})} , \bibinfo{pages}{770--778}.
%Type = Article
\bibitem[{Iancu et~al.(2021)Iancu, Soloviev, Zelioli and Lilius}]{iancuABOshipsInshoreOffshore2021}
\bibinfo{author}{Iancu, B.}, \bibinfo{author}{Soloviev, V.}, \bibinfo{author}{Zelioli, L.}, \bibinfo{author}{Lilius, J.}, \bibinfo{year}{2021}.
\newblock \bibinfo{title}{Aboships---{An} {Inshore} and {Offshore} {Maritime} {Vessel} {Detection} {Dataset} with {Precise} {Annotations}}.
\newblock \bibinfo{journal}{Remote Sensing} \bibinfo{volume}{13}, \bibinfo{pages}{988}.
%Type = Article
\bibitem[{Khanam and Hussain(2024)}]{khanamYOLOv11OverviewKey2024}
\bibinfo{author}{Khanam, R.}, \bibinfo{author}{Hussain, M.}, \bibinfo{year}{2024}.
\newblock \bibinfo{title}{Yolov11: An {Overview} of the {Key} {Architectural} {Enhancements}}.
\newblock \bibinfo{journal}{arXiv preprint} \bibinfo{note}{ArXiv preprint arXiv:2410.17725}.
%Type = Article
\bibitem[{Lin et~al.(2017)Lin, Doll{\' a}r, Girshick, He, Hariharan and Belongie}]{linFeaturePyramidNetworks2017}
\bibinfo{author}{Lin, T.Y.}, \bibinfo{author}{Doll{\' a}r, P.}, \bibinfo{author}{Girshick, R.}, \bibinfo{author}{He, K.}, \bibinfo{author}{Hariharan, B.}, \bibinfo{author}{Belongie, S.}, \bibinfo{year}{2017}.
\newblock \bibinfo{title}{Feature {Pyramid} {Networks} for {Object} {Detection}}.
\newblock \bibinfo{journal}{Proceedings of the {IEEE} {Conference} on {Computer} {Vision} and {Pattern} {Recognition}} , \bibinfo{pages}{936--944}.
%Type = Article
\bibitem[{Lin et~al.(2020)Lin, Goyal, Girshick, He and Doll{\' a}r}]{linFocalLossDense2020}
\bibinfo{author}{Lin, T.Y.}, \bibinfo{author}{Goyal, P.}, \bibinfo{author}{Girshick, R.}, \bibinfo{author}{He, K.}, \bibinfo{author}{Doll{\' a}r, P.}, \bibinfo{year}{2020}.
\newblock \bibinfo{title}{Focal {Loss} for {Dense} {Object} {Detection}}.
\newblock \bibinfo{journal}{IEEE Transactions on Pattern Analysis and Machine Intelligence} \bibinfo{volume}{42}, \bibinfo{pages}{318--327}.
%Type = Article
\bibitem[{Lin et~al.(2014)Lin, Maire, Belongie, Hays, Perona, Ramanan, Doll{\' a}r and Zitnick}]{linMicrosoftCOCOCommon2014}
\bibinfo{author}{Lin, T.Y.}, \bibinfo{author}{Maire, M.}, \bibinfo{author}{Belongie, S.}, \bibinfo{author}{Hays, J.}, \bibinfo{author}{Perona, P.}, \bibinfo{author}{Ramanan, D.}, \bibinfo{author}{Doll{\' a}r, P.}, \bibinfo{author}{Zitnick, C.L.}, \bibinfo{year}{2014}.
\newblock \bibinfo{title}{Microsoft {COCO}: Common {Objects} in {Context}}.
\newblock \bibinfo{journal}{European {Conference} on {Computer} {Vision}} , \bibinfo{pages}{740--755}.
%Type = Article
\bibitem[{Liu et~al.(2024)Liu, Lu, Guo, Ren, Zhu and Lv}]{liuAiOENetAllinOneLowVisibility2024}
\bibinfo{author}{Liu, R.W.}, \bibinfo{author}{Lu, Y.}, \bibinfo{author}{Guo, Y.}, \bibinfo{author}{Ren, W.}, \bibinfo{author}{Zhu, F.}, \bibinfo{author}{Lv, Y.}, \bibinfo{year}{2024}.
\newblock \bibinfo{title}{Aioenet: All-in-{One} {Low}-{Visibility} {Enhancement} to {Improve} {Visual} {Perception} for {Intelligent} {Marine} {Vehicles} {Under} {Severe} {Weather} {Conditions}}.
\newblock \bibinfo{journal}{IEEE Transactions on Intelligent Vehicles} \bibinfo{volume}{9}, \bibinfo{pages}{3811--3826}.
%Type = Article
\bibitem[{Liu et~al.(2016)Liu, Anguelov, Erhan, Szegedy, Reed, Fu and Berg}]{liuSSDSingleShot2016}
\bibinfo{author}{Liu, W.}, \bibinfo{author}{Anguelov, D.}, \bibinfo{author}{Erhan, D.}, \bibinfo{author}{Szegedy, C.}, \bibinfo{author}{Reed, S.}, \bibinfo{author}{Fu, C.Y.}, \bibinfo{author}{Berg, A.C.}, \bibinfo{year}{2016}.
\newblock \bibinfo{title}{Ssd: Single {Shot} {MultiBox} {Detector}}.
\newblock \bibinfo{journal}{European {Conference} on {Computer} {Vision}} , \bibinfo{pages}{21--37}.
%Type = Article
\bibitem[{Lowe(2004)}]{loweDistinctiveImageFeatures2004}
\bibinfo{author}{Lowe, D.G.}, \bibinfo{year}{2004}.
\newblock \bibinfo{title}{Distinctive {Image} {Features} from {Scale}-invariant {Keypoints}}.
\newblock \bibinfo{journal}{International Journal of Computer Vision} \bibinfo{volume}{60}, \bibinfo{pages}{91--110}.
%Type = Article
\bibitem[{Meng et~al.(2023)Meng, Liu, Fan and Fan}]{mengYOLOv5sFogImprovedModel2023}
\bibinfo{author}{Meng, X.}, \bibinfo{author}{Liu, Y.}, \bibinfo{author}{Fan, L.}, \bibinfo{author}{Fan, J.}, \bibinfo{year}{2023}.
\newblock \bibinfo{title}{Yolov5s-{Fog}: An {Improved} {Model} {Based} on {YOLOv5s} for {Object} {Detection} in {Foggy} {Weather} {Scenarios}}.
\newblock \bibinfo{journal}{Sensors} \bibinfo{volume}{23}, \bibinfo{pages}{5321}.
%Type = Article
\bibitem[{Pan et~al.(2020)Pan, Xu, Shi, Zhang, Luo and Lan}]{panDSSNetSimpleDilated2020}
\bibinfo{author}{Pan, B.}, \bibinfo{author}{Xu, X.}, \bibinfo{author}{Shi, Z.}, \bibinfo{author}{Zhang, N.}, \bibinfo{author}{Luo, H.}, \bibinfo{author}{Lan, X.}, \bibinfo{year}{2020}.
\newblock \bibinfo{title}{Dssnet: A {Simple} {Dilated} {Semantic} {Segmentation} {Network} for {Hyperspectral} {Imagery} {Classification}}.
\newblock \bibinfo{journal}{IEEE Geoscience and Remote Sensing Letters} \bibinfo{volume}{17}, \bibinfo{pages}{1968--1972}.
%Type = Article
\bibitem[{Pu et~al.(2023)Pu, Liang, Hao, Yuan, Yang, Zhang, Hu and Huang}]{puRankDETRHighQuality2023}
\bibinfo{author}{Pu, Y.}, \bibinfo{author}{Liang, W.}, \bibinfo{author}{Hao, Y.}, \bibinfo{author}{Yuan, Y.}, \bibinfo{author}{Yang, Y.}, \bibinfo{author}{Zhang, C.}, \bibinfo{author}{Hu, H.}, \bibinfo{author}{Huang, G.}, \bibinfo{year}{2023}.
\newblock \bibinfo{title}{Rank-{DETR} for {High} {Quality} {Object} {Detection}}.
\newblock \bibinfo{journal}{Advances in Neural Information Processing Systems} \bibinfo{volume}{36}, \bibinfo{pages}{16100--16113}.
%Type = Article
\bibitem[{Qi et~al.(2023)Qi, He, Qi, Zhang and Yang}]{qiDynamicSnakeConvolution2023}
\bibinfo{author}{Qi, Y.}, \bibinfo{author}{He, Y.}, \bibinfo{author}{Qi, X.}, \bibinfo{author}{Zhang, Y.}, \bibinfo{author}{Yang, G.}, \bibinfo{year}{2023}.
\newblock \bibinfo{title}{Dynamic {Snake} {Convolution} based on {Topological} {Geometric} {Constraints} for {Tubular} {Structure} {Segmentation}}.
\newblock \bibinfo{journal}{2023 {IEEE}/{CVF} {International} {Conference} on {Computer} {Vision} ({ICCV})} , \bibinfo{pages}{6047--6056}.
%Type = Article
\bibitem[{Redmon et~al.(2016)Redmon, Divvala, Girshick and Farhadi}]{redmonYouOnlyLook2016}
\bibinfo{author}{Redmon, J.}, \bibinfo{author}{Divvala, S.}, \bibinfo{author}{Girshick, R.}, \bibinfo{author}{Farhadi, A.}, \bibinfo{year}{2016}.
\newblock \bibinfo{title}{You {Only} {Look} {Once}: Unified, {Real}-{Time} {Object} {Detection}}.
\newblock \bibinfo{journal}{Proceedings of the {IEEE} {Conference} on {Computer} {Vision} and {Pattern} {Recognition}} , \bibinfo{pages}{779--788}.
%Type = Article
\bibitem[{Redmon and Farhadi(2018)}]{redmonYOLOv3IncrementalImprovement2018}
\bibinfo{author}{Redmon, J.}, \bibinfo{author}{Farhadi, A.}, \bibinfo{year}{2018}.
\newblock \bibinfo{title}{Yolov3: An {Incremental} {Improvement}}.
\newblock \bibinfo{journal}{arXiv preprint} \bibinfo{note}{ArXiv:1804.02767}.
%Type = Article
\bibitem[{Rekavandi et~al.(2023)Rekavandi, Rashidi, Boussaid, Hoefs, Akbas and bennamoun}]{rekavandiTransformersSmallObject2023}
\bibinfo{author}{Rekavandi, A.M.}, \bibinfo{author}{Rashidi, S.}, \bibinfo{author}{Boussaid, F.}, \bibinfo{author}{Hoefs, S.}, \bibinfo{author}{Akbas, E.}, \bibinfo{author}{bennamoun, M.}, \bibinfo{year}{2023}.
\newblock \bibinfo{title}{Transformers in {Small} {Object} {Detection}: A {Benchmark} and {Survey} of {State}-of-the-{Art}}.
\newblock \bibinfo{journal}{arXiv preprint} \bibinfo{note}{ArXiv:2309.04902}.
%Type = Article
\bibitem[{Ren et~al.(2017)Ren, He, Girshick and Sun}]{renFasterRCNNRealtime2017}
\bibinfo{author}{Ren, S.}, \bibinfo{author}{He, K.}, \bibinfo{author}{Girshick, R.}, \bibinfo{author}{Sun, J.}, \bibinfo{year}{2017}.
\newblock \bibinfo{title}{Faster {R}-{CNN}: Towards {Real}-time {Object} {Detection} with {Region} {Proposal} {Networks}}.
\newblock \bibinfo{journal}{IEEE Transactions on Pattern Analysis and Machine Intelligence} \bibinfo{volume}{39}, \bibinfo{pages}{1137--1149}.
%Type = Article
\bibitem[{Shao et~al.(2018)Shao, Wu, Wang, Du and Li}]{shaoSeaShipsLargeScalePrecisely2018}
\bibinfo{author}{Shao, Z.}, \bibinfo{author}{Wu, W.}, \bibinfo{author}{Wang, Z.}, \bibinfo{author}{Du, W.}, \bibinfo{author}{Li, C.}, \bibinfo{year}{2018}.
\newblock \bibinfo{title}{Seaships: A {Large}-{Scale} {Precisely} {Annotated} {Dataset} for {Ship} {Detection}}.
\newblock \bibinfo{journal}{IEEE Transactions on Multimedia} \bibinfo{volume}{20}, \bibinfo{pages}{2593--2604}.
%Type = Article
\bibitem[{Terven and Cordova-Esparza(2023)}]{tervenComprehensiveReviewYOLO2023}
\bibinfo{author}{Terven, J.}, \bibinfo{author}{Cordova-Esparza, D.}, \bibinfo{year}{2023}.
\newblock \bibinfo{title}{A {Comprehensive} {Review} of {YOLO} {Architectures} in {Computer} {Vision}: From {YOLOv1} to {YOLOv8} and {YOLO}-{NAS}}.
\newblock \bibinfo{journal}{Machine Learning and Knowledge Extraction} \bibinfo{volume}{5}, \bibinfo{pages}{1680--1716}.
\newblock \bibinfo{note}{ArXiv preprint arXiv:2304.00501}.
%Type = Article
\bibitem[{Vaswani et~al.(2017)Vaswani, Shazeer, Parmar, Uszkoreit, Jones, Gomez, Kaiser and Polosukhin}]{vaswaniAttentionAllYou2017a}
\bibinfo{author}{Vaswani, A.}, \bibinfo{author}{Shazeer, N.}, \bibinfo{author}{Parmar, N.}, \bibinfo{author}{Uszkoreit, J.}, \bibinfo{author}{Jones, L.}, \bibinfo{author}{Gomez, A.N.}, \bibinfo{author}{Kaiser, L.}, \bibinfo{author}{Polosukhin, I.}, \bibinfo{year}{2017}.
\newblock \bibinfo{title}{Attention is {All} you {Need}}.
\newblock \bibinfo{journal}{International Conference on Neural Information Processing Systems} , \bibinfo{pages}{6000--6010}.
%Type = Article
\bibitem[{Wang et~al.(2023a)Wang, He, Nie, Guo, Liu, Han and Wang}]{wangGoldYOLOEfficientObject2023}
\bibinfo{author}{Wang, C.}, \bibinfo{author}{He, W.}, \bibinfo{author}{Nie, Y.}, \bibinfo{author}{Guo, J.}, \bibinfo{author}{Liu, C.}, \bibinfo{author}{Han, K.}, \bibinfo{author}{Wang, Y.}, \bibinfo{year}{2023}a.
\newblock \bibinfo{title}{Gold-{YOLO}: Efficient {Object} {Detector} via {Gather}-and-{Distribute} {Mechanism}}.
\newblock \bibinfo{journal}{International {Conference} on {Neural} {Information} {Processing} {Systems}} , \bibinfo{pages}{51094--51112}.
%Type = Article
\bibitem[{Wang et~al.(2023b)Wang, Bochkovskiy and Liao}]{wangYOLOv7TrainableBagofFreebies2023}
\bibinfo{author}{Wang, C.Y.}, \bibinfo{author}{Bochkovskiy, A.}, \bibinfo{author}{Liao, H.Y.M.}, \bibinfo{year}{2023}b.
\newblock \bibinfo{title}{Yolov7: Trainable {Bag}-of-{Freebies} {Sets} {New} {State}-of-the-{Art} for {Real}-{Time} {Object} {Detectors}}.
\newblock \bibinfo{journal}{Proceedings of the {IEEE} {Conference} on {Computer} {Vision} and {Pattern} {Recognition}} , \bibinfo{pages}{7464--7475}.
%Type = Article
\bibitem[{Wang et~al.(2024)Wang, Wang, Wei, Han and Feng}]{wangMarineVesselDetection2024}
\bibinfo{author}{Wang, N.}, \bibinfo{author}{Wang, Y.}, \bibinfo{author}{Wei, Y.}, \bibinfo{author}{Han, B.}, \bibinfo{author}{Feng, Y.}, \bibinfo{year}{2024}.
\newblock \bibinfo{title}{Marine {Vessel} {Detection} {Dataset} and {Benchmark} for {Unmanned} {Surface} {Vehicles}}.
\newblock \bibinfo{journal}{Applied Ocean Research} \bibinfo{volume}{142}, \bibinfo{pages}{103835}.
%Type = Article
\bibitem[{Wang et~al.(2021)Wang, Xie, Li, Fan, Song, Liang, Lu, Luo and Shao}]{wangPyramidVisionTransformer2021}
\bibinfo{author}{Wang, W.}, \bibinfo{author}{Xie, E.}, \bibinfo{author}{Li, X.}, \bibinfo{author}{Fan, D.P.}, \bibinfo{author}{Song, K.}, \bibinfo{author}{Liang, D.}, \bibinfo{author}{Lu, T.}, \bibinfo{author}{Luo, P.}, \bibinfo{author}{Shao, L.}, \bibinfo{year}{2021}.
\newblock \bibinfo{title}{Pyramid {Vision} {Transformer}: A {Versatile} {Backbone} for {Dense} {Prediction} without {Convolutions}}.
\newblock \bibinfo{journal}{Proceedings of the {IEEE} {International} {Conference} on {Computer} {Vision}} , \bibinfo{pages}{548--558}.
%Type = Article
\bibitem[{Wang and Leuven(2024)}]{wangNavigatingWatersObject2024}
\bibinfo{author}{Wang, Y.}, \bibinfo{author}{Leuven, K.}, \bibinfo{year}{2024}.
\newblock \bibinfo{title}{Navigating the {Waters} of {Object} {Detection}: Evaluating the {Robustness} of {Real}-time {Object} {Detection} {Models} for {Autonomous} {Surface} {Vehicles}}.
\newblock \bibinfo{journal}{Proceedings of the {IEEE} {Conference} on {Artificial} {Intelligence}} , \bibinfo{pages}{993--1000}.
%Type = Article
\bibitem[{Wei et~al.(2022)Wei, Liu, Xu, Dai, Dai and Xu}]{weiDWRSegRethinkingEfficient2022}
\bibinfo{author}{Wei, H.}, \bibinfo{author}{Liu, X.}, \bibinfo{author}{Xu, S.}, \bibinfo{author}{Dai, Z.}, \bibinfo{author}{Dai, Y.}, \bibinfo{author}{Xu, X.}, \bibinfo{year}{2022}.
\newblock \bibinfo{title}{Dwrseg: Rethinking {Efficient} {Acquisition} of {Multi}-scale {Contextual} {Information} for {Real}-time {Semantic} {Segmentation}}.
\newblock \bibinfo{journal}{arXiv preprint} \bibinfo{note}{ArXiv:2212.01173}.
%Type = Article
\bibitem[{Xia et~al.(2022)Xia, Pan, Song, Li and Huang}]{xiaVisionTransformerDeformable2022}
\bibinfo{author}{Xia, Z.}, \bibinfo{author}{Pan, X.}, \bibinfo{author}{Song, S.}, \bibinfo{author}{Li, L.E.}, \bibinfo{author}{Huang, G.}, \bibinfo{year}{2022}.
\newblock \bibinfo{title}{Vision {Transformer} {With} {Deformable} {Attention}}.
\newblock \bibinfo{journal}{Proceedings of the {IEEE} {Conference} on {Computer} {Vision} and {Pattern} {Recognition}} , \bibinfo{pages}{4794--4803}.
%Type = Article
\bibitem[{Xing et~al.(2023)Xing, Ren, Fan and Zhang}]{xingSDETRTransformerModel2023}
\bibinfo{author}{Xing, Z.}, \bibinfo{author}{Ren, J.}, \bibinfo{author}{Fan, X.}, \bibinfo{author}{Zhang, Y.}, \bibinfo{year}{2023}.
\newblock \bibinfo{title}{S-{DETR}: A {Transformer} {Model} for {Real}-{Time} {Detection} of {Marine} {Ships}}.
\newblock \bibinfo{journal}{Journal of Marine Science and Engineering} \bibinfo{volume}{11}, \bibinfo{pages}{696}.
%Type = Article
\bibitem[{Xu et~al.(2021)Xu, Zhang, Hu, Wang, Wang, Wei, Bai and Liu}]{xuEndtoEndSemiSupervisedObject2021}
\bibinfo{author}{Xu, M.}, \bibinfo{author}{Zhang, Z.}, \bibinfo{author}{Hu, H.}, \bibinfo{author}{Wang, J.}, \bibinfo{author}{Wang, L.}, \bibinfo{author}{Wei, F.}, \bibinfo{author}{Bai, X.}, \bibinfo{author}{Liu, Z.}, \bibinfo{year}{2021}.
\newblock \bibinfo{title}{End-to-{End} {Semi}-{Supervised} {Object} {Detection} with {Soft} {Teacher}}.
\newblock \bibinfo{journal}{International {Conference} on {Computer} {Vision} ({ICCV})} , \bibinfo{pages}{3040--3049}.
%Type = Article
\bibitem[{Yang et~al.(2024)Yang, Zhang, Wang and Liu}]{yangLightweightTheorydrivenNetwork2024}
\bibinfo{author}{Yang, Z.}, \bibinfo{author}{Zhang, P.}, \bibinfo{author}{Wang, N.}, \bibinfo{author}{Liu, T.}, \bibinfo{year}{2024}.
\newblock \bibinfo{title}{A {Lightweight} {Theory}-driven {Network} and {Its} {Validation} on {Public} {Fully} {Polarized} {Ship} {Detection} {Dataset}}.
\newblock \bibinfo{journal}{IEEE Journal of Selected Topics in Applied Earth Observations and Remote Sensing} \bibinfo{volume}{17}, \bibinfo{pages}{3755--3767}.
%Type = Article
\bibitem[{Ye et~al.(2023)Ye, Ke, Li, Tai, Tang, Danelljan and Yu}]{yeCascadeDETRDelvingHighQuality2023}
\bibinfo{author}{Ye, M.}, \bibinfo{author}{Ke, L.}, \bibinfo{author}{Li, S.}, \bibinfo{author}{Tai, Y.W.}, \bibinfo{author}{Tang, C.K.}, \bibinfo{author}{Danelljan, M.}, \bibinfo{author}{Yu, F.}, \bibinfo{year}{2023}.
\newblock \bibinfo{title}{Cascade-{DETR}: Delving into {High}-{Quality} {Universal} {Object} {Detection}}.
\newblock \bibinfo{journal}{Proceedings of the {IEEE} {International} {Conference} on {Computer} {Vision}} , \bibinfo{pages}{6704--6714}.
%Type = Article
\bibitem[{Yu and Koltun(2015)}]{yuMultiscaleContextAggregation2016}
\bibinfo{author}{Yu, F.}, \bibinfo{author}{Koltun, V.}, \bibinfo{year}{2015}.
\newblock \bibinfo{title}{Multi-scale {Context} {Aggregation} by {Dilated} {Convolutions}}.
\newblock \bibinfo{journal}{arXiv preprint} \bibinfo{note}{ArXiv:2410.17725}.
%Type = Article
\bibitem[{Zhang et~al.(2022)Zhang, Er, Gao and Wu}]{zhangHighPerformanceShip2022}
\bibinfo{author}{Zhang, Y.}, \bibinfo{author}{Er, M.J.}, \bibinfo{author}{Gao, W.}, \bibinfo{author}{Wu, J.}, \bibinfo{year}{2022}.
\newblock \bibinfo{title}{High {Performance} {Ship} {Detection} via {Transformer} and {Feature} {Distillation}}.
\newblock \bibinfo{journal}{2022 5th {International} {Conference} on {Intelligent} {Autonomous} {Systems} ({ICoIAS})} , \bibinfo{pages}{31--36}.
%Type = Article
\bibitem[{Zhang et~al.(2025)Zhang, Wu, Liu and Peng}]{zhangCPAEnhancerChainofThoughtPrompted2025}
\bibinfo{author}{Zhang, Y.}, \bibinfo{author}{Wu, Y.}, \bibinfo{author}{Liu, Y.}, \bibinfo{author}{Peng, X.}, \bibinfo{year}{2025}.
\newblock \bibinfo{title}{Cpa-{Enhancer}: Chain-of-{Thought} {Prompted} {Adaptive} {Enhancer} for {Downstream} {Vision} {Tasks} {Under} {Unknown} {Degradations}}.
\newblock \bibinfo{journal}{IEEE {International} {Conference} on {Acoustics}, {Speech} and {Signal} {Processing} ({ICASSP})} , \bibinfo{pages}{1--5}.
%Type = Article
\bibitem[{Zheng and Zhang(2020)}]{zhengMcshipsLargescaleShip2020}
\bibinfo{author}{Zheng, Y.}, \bibinfo{author}{Zhang, S.}, \bibinfo{year}{2020}.
\newblock \bibinfo{title}{Mcships: A {Large}-scale {Ship} {Dataset} for {Detection} and {Fine}-grained {Categorization} in the {Wild}}.
\newblock \bibinfo{journal}{Proceedings of the {IEEE} {International} {Conference} on {Multimedia} and {Expo}} , \bibinfo{pages}{1--6}.
%Type = Article
\bibitem[{Zhu et~al.(2019)Zhu, Hu, Lin and Dai}]{zhuDeformableConvNetsV22019}
\bibinfo{author}{Zhu, X.}, \bibinfo{author}{Hu, H.}, \bibinfo{author}{Lin, S.}, \bibinfo{author}{Dai, J.}, \bibinfo{year}{2019}.
\newblock \bibinfo{title}{Deformable {ConvNets} {V2}: More {Deformable}, {Better} {Results}}.
\newblock \bibinfo{journal}{Proceedings of the {IEEE} {Conference} on {Computer} {Vision} and {Pattern} {Recognition}} , \bibinfo{pages}{9300--9308}.
%Type = Article
\bibitem[{Zhu et~al.(2021)Zhu, Su, Lu, Li, Wang and Dai}]{zhuDeformableDetrDeformable2021}
\bibinfo{author}{Zhu, X.}, \bibinfo{author}{Su, W.}, \bibinfo{author}{Lu, L.}, \bibinfo{author}{Li, B.}, \bibinfo{author}{Wang, X.}, \bibinfo{author}{Dai, J.}, \bibinfo{year}{2021}.
\newblock \bibinfo{title}{Deformable {Detr}: Deformable {Transformers} for {End}-to-{End} {Object} {Detection}}.
\newblock \bibinfo{journal}{International {Conference} on {Learning} {Representations}} , \bibinfo{pages}{1041--1056}.

\end{thebibliography}


%\begin{thebibliography}{00}



%\end{thebibliography}

\end{document}